\pdfoutput=1

\documentclass[11pt]{article}
\usepackage[dvipsnames]{xcolor}

\usepackage[final]{acl}

\usepackage{times}
\usepackage{latexsym}
\usepackage{multirow}
\usepackage{todonotes}
\usepackage{comment}
\usepackage[multiple]{footmisc}
\usepackage{fancyvrb} 
\usepackage[most]{tcolorbox}
\usepackage{enumitem}
\usepackage{algpseudocode}
\usepackage{comment}
\usepackage{soul}
\usepackage{CJKutf8}
\usepackage[T1]{fontenc}

\usepackage[utf8]{inputenc}

\usepackage{microtype}

\usepackage{inconsolata}
\usepackage{algorithm}
\usepackage{algpseudocode}

\usepackage{pifont}
\usepackage[normalem]{ulem}
\usepackage{amsmath}
\usepackage{multirow}
\usepackage{multirow}
\usepackage[normalem]{ulem}

\usepackage{graphicx}

\title{Iterative Repair with Weak Verifiers \\ for Few-shot Transfer in KBQA with Unanswerability}
\author{
Riya Sawhney$^{\ddag,\S}$, 
Samrat Yadav$^\ddag$, 
Indrajit Bhattacharya$^\dag$, Mausam$^\ddag$\\
$^\ddag$Indian Institute of Technology, Delhi, 
$^\S$Graviton Research Capital,
$^\dag$KnowDis AI\\
\texttt{
{\small
riya.sawhney@outlook.com, samratya23@gmail.com,
indrajitb@gmail.com,  mausam@cse.iitd.ac.in
}
}
}

\begin{document}
\maketitle

\newcommand{\sysold}{FuSIC-KBQA}
\newcommand{\sys}{FUn-FuSIC}
\newenvironment{algocolor}{%
   \setlength{\parindent}{0pt}
   \itshape
   \color{blue}
}{}

\begin{abstract}
Real-world applications of KBQA require models to detect different types of unanswerable questions with a limited volume of in-domain labeled training data. We propose the novel task of few-shot transfer for KBQA with unanswerable questions. The state-of-the-art KBQA few-shot transfer model (FuSIC-KBQA) uses an iterative repair strategy that assumes that all questions are answerable. As a remedy, we present FUn-FuSIC -- a novel solution for our task that extends FuSIC-KBQA with Feedback for Unanswerability (FUn), which is an iterative repair strategy for answerable as well as unanswerable questions. FUn uses feedback from a suite of strong and weak verifiers, and an adaptation of self-consistency for unanswerability for assessing answerability of questions. Our experiments show that FUn-FuSIC significantly outperforms suitable adaptations of multiple LLM-based and supervised SoTA models on our task, while establishing a new SoTA performance for answerable few-shot transfer as well. We have made datasets and other resources publicly available\footnote{\url{https://github.com/dair-iitd/FUn-FuSIC}}
\end{abstract}

\section{Introduction}\label{sec:intro}
The semantic parsing formulation of the Knowledge Base Question Answering (KBQA) task takes as input a Knowledge Base (KB) and a natural language question, and outputs a logical form (or program) that produces the answer upon execution over the KB. 
KBQA has important real-world applications, which require KBQA systems to be low-resource (i.e., trained only with a few task-specific labeled examples), and robust, specifically able to identify questions that cannot be answered from the KB. 

Traditional supervised models (e.g., \cite{ye:acl2022rngkbqa,shu-etal-2022-tiara,gu-etal-2023-dont}) and even recent LLM few-shot in-context learning (FS-ICL) architectures~\cite{li:acl2023kbbinder,nie:arxiv2023kbcoder} for KBQA fall short in both aspects. 
Limited recent work has addressed these independently --  in-domain methods for KBQA with unanswerability trained with large labeled data~\cite{patidar-etal-2023-knowledge, faldu:acl2024}, and \sysold{} for few-shot transfer assuming answerable questions \cite{patidar-etal-2024-fusic}. 
No existing single KBQA model simultaneously addresses both desiderata.

In response, we propose the novel task of {\em few-shot transfer learning for KBQA with unanswerability}.
Specifically, the target domain has only a few labeled examples of answerable and unanswerable questions, while the source domain has thousands of labeled examples, but containing {\em only} answerable questions. 

For few-shot KBQA transfer, \sysold{} uses a retrieve-then-generate framework: retrieval of relevant schema and KB snippets followed by  an LLM-based generation and a subsequent iterative execution-error-guided repair. Specifically, multiple feedback-guided repair iterations are executed, checking emptiness of answers obtained by executing the generated program as indication of correctness, until a \emph{non-empty} answer is obtained. This naturally fails when questions are allowed to be unanswerable.

A simple-fix for addressing {\bf U}nanswerability (\sysold{}-U) is to drop the inappropriate repair step, and modify the LLM prompt to accommodate unanswerable questions, along with relevant in-context exemplars. 
Unlike in studies for unanswerability in general QA~\cite{slobodkin-etal-2023-curious}, we found that \sysold{}-U mostly generates incorrect logical forms for unanswerable questions.

As a remedy, we design a novel solution: \sys{} ({\bf F}eedback for {\bf Un}answerability in \sysold{}). The key idea is to {\em modify} iterative repair, which earlier relied on a single strong verifier for the logical form's incorrectness, 
to rely on a {\em suite of strong and weak verifiers}, where strong verifiers identify certain errors, whereas weak verifiers identify potential errors in the current logical form. 

\sys{}'s verifiers consider both the logical form and the answer. 
For answers, non-emptiness check is now a weak verifier, given potentially unanswerable questions.
For logical forms, we use strong verifiers to identify obvious syntactic and semantic errors.
We also propose a novel verifier involving a 3-component LLM-based pipeline: non-equivalence of the original question and the back-translation of the logical form. 
This verifier is also weak, due to potential errors in back-translation as well as in equivalence classification. 
Using such iterative strong and weak verification based repair, \sys{} constructs a set of candidate logical forms. 
For selecting the consensus logical form from this set, using the majority answer as in self-consistency~\cite{DBLP:conf/iclr/0002WSLCNCZ23} breaks down in the face of unanswerability.
We introduce {\em self-consistency for unanswerability}, which assesses the {\em likelihood} of the majority answer, empty or otherwise, to select the consensus logical form.

Since no datasets exist for our novel task, we create two new datasets for KBQA transfer with unanswerable questions. 
Our experiments show that
\sys{} comprehensively outperforms different categories of SoTA models suitably adapted for this task, including LLM-based and more traditional models.
We further find that iterative repair of logical forms using weak verifiers holds promise for even for KBQA with only answerable questions. 
Using experiments over benchmark datasets for this task, we show that the restriction of \sys{} for the answerable setting improves upon the SoTA model for the task.

In summary, our specific contributions are as follows.
{\bf (a)} We propose the problem of few-shot transfer for KBQA with unanswerability. {\bf (b)} We present \sys{} 
that uses iterative repair with error feedback from a diverse suite of strong and weak verifiers.
{\bf (c)} We create new datasets for the proposed task, which we make public. 
{\bf (d)} We show that \sys{} outperforms adaptations of  SoTA KBQA models for this new task. 
{\bf (e)} We also show that even for answerable-only KBQA, \sys{} outperforms the corresponding SoTA model.

\section{Related Work}\label{sec:rw}
In-domain KBQA using supervised models
~\cite{saxena:acl2022,zhang:acl2022subgraph,mitra:naacl2022cmhopkgqa,wang:naacl2022mhopkgqa,das:icml2022subgraphcbr,ye:acl2022rngkbqa,chen:acl2021retrack,das:emnlp2021cbr,shu-etal-2022-tiara,gu-etal-2023-dont} and using LLM few-shot approaches~\cite{li:acl2023kbbinder,nie:arxiv2023kbcoder,shu:arxiv2023bottlenecks} is well explored in literature.
These use high volumes of labeled data, either for training or selecting the most relevant few shot exemplars.

For in-domain KBQA, unanswerability has recently been studied~\cite{patidar-etal-2023-knowledge,faldu:acl2024}.
~\newcite{patidar-etal-2023-knowledge} create the GrailQAbility dataset with different categories of unanswerability, and show the inadequacy of superficial adaptations of answerable-only KBQA models.
RetinaQA~\cite{faldu:acl2024} is the SoTA model for KBQA unanswerability.
However, this also requires large volumes of training data.

For KBQA transfer~\cite{cao:acl2022progxfer,ravishankar:emnlp2022}, 
low-resource was originally not a focus.
More recently, few-shot transfer for KBQA has been addressed by \sysold{}~\cite{patidar-etal-2024-fusic}.
\sysold{} uses a retrieve-then-generate framework with an LLM-based generation stage with iterative error feedback based repair.
However, this formulation assumes answerability of all questions.

Simple LLM prompting techniques have been used to address unanswerability outside of KBQA~\cite{slobodkin-etal-2023-curious}, but without any notion of feedback or iterative repair.
Other approaches ~\cite{shinn:neurips2023reflexion,chen2023teaching} use execution based refinement for program generation but without any notion of unanswerability. 

FUn's iterative repair idea may be useful in other natural language to program generation tasks where non-existence of a program with the required specification has not been studied to the best of our knowledge, such as NL-to-SQL ~\cite{dong2023c3zeroshottexttosqlchatgpt,pourreza2023dinsqldecomposedincontextlearning} and program self-repair using LLMs~\cite{olausson:iclr2024,madaan2023selfrefine,grattafiori2024llama3herdmodels}.
\section{Background \& Problem Definition}\label{sec:problem_and_approach}

\begin{figure*}[h]
\centering
\includegraphics[width=1.00\textwidth]{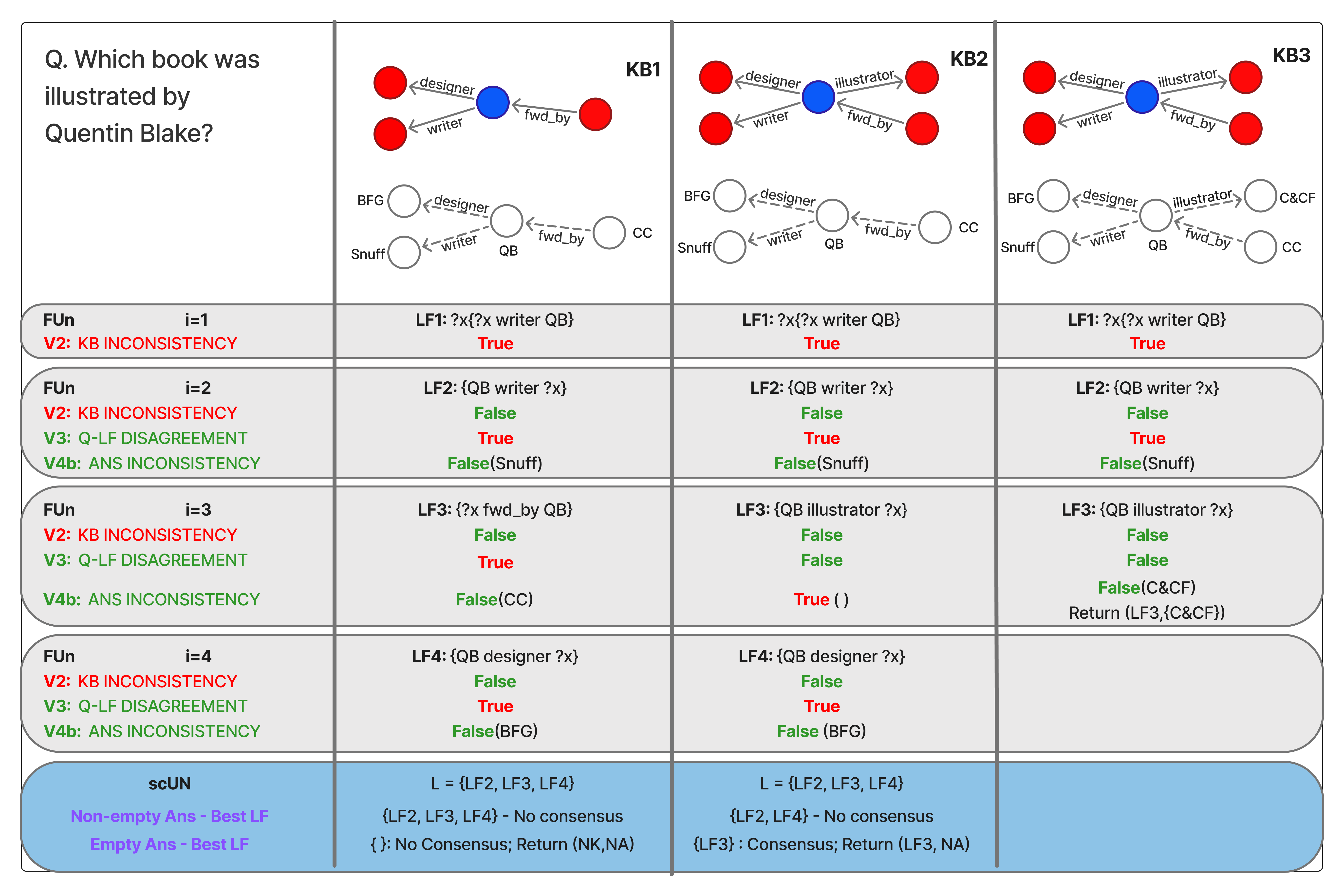}
\caption{{\bf F}eedback with {\bf Un}answerability (FUn) and {\bf s}elf {\bf c}onsistency for {\bf Un}answerability (scUn) for a question when executed over different KBs with $\leq4$ iterations. 
The question is {\bf answerable} for KB3, but {\bf unanswerable} for KB1 ({\bf schema incompleteness}) and KB2 ({\bf data incompleteness}). 
In the KB depictions, the top graph represents the {\bf schema} with different node colors for different {\bf entity types}, and the bottom graph represents the {\bf data}. 
(Names of different (real) books related to the author entity in the question are abbreviated in the data graph.)
{\bf FUn} iterations are shown in the \colorbox{Gray}{gray blocks}, with {\bf Strong Verifiers} named using \textcolor{BrickRed}{red text} and {\bf Weak Verifiers} using \textcolor{ForestGreen}{green text} and $i$ denotes the iteration number. 
The outcome of verification is denoted as \textcolor{BrickRed}{{\bf True}} or \textcolor{ForestGreen}{{\bf False}}, and the non-empty answer is shown when V4b returns \textcolor{ForestGreen}{{\bf False}}.
The Syntax Error Verifier (V1) is omitted for brevity. 
{\bf scUn} is shown in the \colorbox{CornflowerBlue}{blue blocks}.
The candidate logical forms ($L$) for sCun are shown at the top.
For \textcolor{RoyalPurple}{{\bf Non-empty Ans}} and \textcolor{RoyalPurple}{{\bf Empty Ans}} agreement checks for, the outcome is either `Consensus' or `No consensus'. 
At the end, \sys{} `Return's a logical form (possibly \texttt{NK}) and an answer (possibly \texttt{NA}).
}
\label{fig:FUn}
\end{figure*}


A Knowledge Base (KB) $G$ consists of a schema and data.
The schema consists of entity types (or classes) $T$ and binary relations $R$ defined over pairs of types. 
The data consists of entities $E$ as instances of types $T$, and triples or facts $F\subseteq E \times R \times E$.
Given a {\em target KB} $G^t$ and a natural language question $q^t$, the {\bf basic KBQA} task is to generate a structured query or logical form $l^t$ (in a KB query language, such as SPARQL), which when executed over $G^t$ returns an answer $A^t$ ($A^t\subset E$ in general).
Other than SPARQL
~\cite{patidar-etal-2024-fusic}, niche languages such as s-expressions ~\cite{li:acl2023kbbinder,gu-etal-2023-dont} are commonly used for logical forms in KBQA.
In {\bf supervised in-domain KBQA}, the target has large volumes of labeled training examples of questions and associated logical forms.
For {\bf few-shot in-domain KBQA} in contrast, {\em target few-shots} $D^t$ contain tens of labeled training examples.
In {\bf few-shot transfer learning for KBQA}~\cite{patidar-etal-2024-fusic}, a related source domain has a {\em source KB} $G^s$ (with its own types, relations, entities and facts), and a larger {\em source training set} $D^s$ with thousands of labeled training examples.

Following \newcite{patidar-etal-2023-knowledge}, a question $q$ is {\bf answerable} for a KB $G$ if it admits a corresponding logical form $l$ which when executed over $G$ returns the ideal {\em non-empty answer} $A$. 
A question is {\bf unanswerable} if it either (a) does not have a valid logical form for $G$ (schema-level unanswerability), or (b) it has a valid logical form $l$ for $G$, but $l$ returns an empty answer upon execution on $G$, different from the ideal non-empty answer (data-level unanswerability). 
Schema-level unanswerability arises due to missing types and relations, 
while missing entities and facts lead to data-level unanswerability.
However, absence in $G$ of {\em any entity mentioned in the question} is categorized as schema level unanswerability, since it invalidates the logical form. 
More details are in Appendix ~\ref{kbqa_unanswerability}.

In {\bf KBQA with unanswerability}~\cite{patidar-etal-2023-knowledge,faldu:acl2024}, given a question $q$, the model needs to output {\bf (a)} a logical form $l$ and a non-empty answer \texttt{A} for answerable $q$, {\bf (b)} $l=$ \texttt{NK} (No Knowledge) for schema-level unanswerable $q$, or {\bf (c)} a valid logical form $l$ and $a=$ \texttt{NA} (No Answer) for data-level unanswerable $q$.
In the supervised in-domain setting, this task involves large volumes of labeled training questions, containing both answerable and unanswerable, for the target.

We now define our problem of interest: {\bf few-shot transfer learning for KBQA with unanswerability}. 
A target question $q^t$ may be answerable or unanswerable due to missing schema or data in the target KB $G^t$. 
Target few-shot examples $D^t$ contain both answerable and unanswerable questions of different categories.
The source training data $D^s$ has large volumes of labeled training data.
Considering real world constraints, where most KBQA datasets contain only answerable questions, we assume that $D^s$ contains {\em only answerable questions}.
Compared to the earlier few-shot KBQA transfer task definition, now there is additionally an unanswerability mismatch between the source and the target distribution. 
More details are in the Appendix (Sec. ~\ref{subsec:appendix_problem_details}).

\section{Proposed Approach: \sys{}}\label{subsec:approach}

Our proposed model \sys{} preserves the basic architecture of \sysold{} and adapts its iterative repair strategy for unanswerability. 
The high-level algorithm is described in Algo.~\ref{algo:fun_fusic}. 
(Since the algorithms are not specific to the transfer task,  we use $q$ instead of $q^t$ for brevity.)
Preserving the retrieve-then-generate framework of \sysold{}, the retrieval stage (line 2) performs KB retrieval for $q^t$ using a set $R$ of one or more supervised retrievers. 
Each retriever $R_i$ is source-trained and further target fine-tuned if required. 
The retrieval output $r$ of each $R_i$ consists of relevant schema elements (types and relations) for $q^t$, and data paths emanating from mentioned entities in $q^t$.
The union of these, along with $q^t$, is fed to the generation stage, which uses prompting with an LLM $\mathcal{L}$ to generate logical forms using the target few-shots $D^t$.
More details of the retrieval stage are in the Appendix (Sec. \ref{fusic_high_level}). 

\begin{algorithm}
\caption{{\small FUn-FuSIC}$(q,G^t,D^t,R,V^s,V^w,\mathcal{L})$}
\label{algo:fun_fusic}
\begin{algorithmic}[1] 
\State $r=\{\}$
\State $\mbox{ {\bf for} }i=1\mbox{ to }k \mbox{ {\bf do} } r=r\bigcup R_i(q,G^t)$
\begin{algocolor}
\State $l=\mbox{PUn}(\mathcal{L}, I, q,r,D^t)$
\State $(e,l,A,L)=\mbox{FUn}(\mathcal{L},q,l,n,V^s,V^w,G^t)$
\end{algocolor}
\State $\mbox{{\bf if} }(e)\mbox{ {\bf return}}(l^*,A^*)$

\State $\mbox{{\bf else} 
 {\bf return}}$ \begin{algocolor}$\mbox{scUn}(q,L,\mathcal{L})$
\end{algocolor}
\end{algorithmic}
\end{algorithm}

\sys{} differs from \sysold{} in its iterative repair strategy in lines 3, 4 and 6 of Algo.~\ref{algo:fun_fusic}.
The LLM generation instruction is modified to admit the possibility of unanswerability, and the few shots are modified to include examples of
unanswerable questions.
However, this simple approach is error-prone.
So, we bias the instruction towards one type of error. 
Specifically, when uncertain about answerability of the question, {\bf P}rompting for {\bf Un}answerability (PUn) (line 3) instructs $\mathcal{L}$ to generate a (possibly incorrect) logical form instead of $l=\texttt{NK}$. 
The detailed prompt $I$ is in the Appendix (Sec.~\ref{pun_prompt}).  

We now come to the more significant modifications.
First, $l^{(0)}$ is iteratively repaired using feedback as before, but this step is adapted for unanswerability.
This iterative repair, which we name {\bf F}eedback for {\bf Un}answerability (FUn) ({\bf line 4}), either confidently outputs a single logical form $l$ (with corresponding answer $A$) (line 5) or generates a set $L$ of candidate logical forms, which is further analyzed for a consensus logical form and answer.
For this, we introduce {\bf s}elf-{\bf c}onsistency for {\bf Un}answerability (scUn) ({\bf line 6}).
scUn assesses the likelihood of the majority answer in $L$, empty or otherwise, to produce the final output.
In the rest of this section, we describe FUn and then scUn.
Fig.~\ref{fig:FUn} illustrates flow of FUn and scUn using examples.
A real example of FUn execution is in Sec.~\ref{subsec_fun_real}.

\begin{algorithm}
\caption{$\mbox{FUn}(q,l,n,V^s,V^w,G,\mathcal{L})$}
\label{algo:fun}
\begin{algorithmic}[1] 
\State $i=0$, $k=0$, $L=\{\}$, $F=\mbox{""}$
\While{$i\leq n$}
    \For{$j=1\mbox{ to }k_1$}
        \State $(e,f)=V^s_j(l^{(i)},q,G)$
        \State $F=\mbox{Append}(F,f)$
        \State $\mbox{{\bf if} }(!e)\mbox{ {\bf break}}$
    \EndFor
    \For{$j=1\mbox{ to }k_2$}
        \State $(e,f)=V^w_j(l^{(i)},q,G)$
        \State $F=\mbox{Append}(F,f)$
        \State $\mbox{{\bf if} }(e)\mbox{ }L=L\bigcup\{l^{(i)}\}$; $k++$
    \EndFor
    \State $i = i+1$
    \State $l^{(i)}=Gen(\mathcal{L}, I,q,F)$
    \State $\mbox{{\bf if} }$ $(k=k_2)$ $\mbox{ {\bf return}}(\mbox{T},l^{(i)},\mbox{X}(l^{(i)},G),L)$
\EndWhile
\State $\mbox{{\bf return}}(\mbox{F},l,\{\},L)$
\end{algorithmic}
\end{algorithm}

The FUn algorithm is described in Algo.~\ref{algo:fun}
Starting with the initial logical form $l^{(0)}$, FUn performs at most $n$ verify-and-repair iterations to create a candidate set $L$ of probable logical forms.
Fig.~\ref{fig:FUn} shows 3 FUn iterations for KB1 and KB2, and 2 for KB3.
In the $i^{th}$ iteration, FUn generates a new logical form $l^{(i)}$ by prompting $\mathcal{L}$ using $q^t$ and feedback $F$ received from checks in all previous iterations (line 13).
$l^{(i)}$ goes through a sequence of verifications.
FUn uses two sets of verifiers.
The {\em strong verifiers} $V^s$ are guaranteed to be correct, while the {\em weak verifiers} $V^w$ are potentially erroneous.
$k_1$ and $k_2$ denote the total number of strong and weak verifiers respectively.
The specific verifiers that we use in this paper are defined later in the section.
A template-based feedback string $f$ is appended to the generation prompt for $l^{(i+1)}$ based on the specific verifier that $l^{(i)}$ failed.  
If $l^{(i)}$ fails a strong verifier, it is rejected (line 6).
In the example, this happens for all three KBs in iteration 1.
If $l^{(i)}$ passes all checks, strong and weak (line 15), FUn terminates by outputting $(l=l^{(i)}, A=A^{(i)})$, where $A^{(i)}$ is the answer obtained by executing $l^{(i)}$ (denoted $\mbox{X}(l^{(i)},G)$).
This happens in iteration 3 for KB3.
Otherwise, if $l^{(i)}$ passes at least one weak verifier but not all, it is added to candidate logical form set $L$ (line 11). 
This happens for iterations 2, 3 and 4 for KB1 and KB2.


\paragraph{Logical form verifiers:}
FUn uses a suite of verifiers, categorized as strong ($V^s$) and weak ($V^w$).
These may be syntactic, semantic, or execution-based, defined using simple rules or complex LLM functions over $l$, $q$ and $G$.
Note that unlike unit tests in program synthesis, the verifiers {\em do not} have knowledge of the gold logical form, the gold answer or answerability of the question. 

We now briefly describe the specific verifiers that we use for this paper.
Additional details about the verifiers are in the Appendix (Sec.~\ref{fun_prompt} and Sec.~\ref{subsec_appendix_error_check}).
Note that FUn is a {\em framework} that is capable of working with a wholly different suite of meaningful verifiers.

\paragraph{(V1) Syntax Error:} As in \sysold{}, this verifier executes the logical form $l$ over $G$ and
checks for syntax error.
This is a strong check --- a valid logical form cannot have syntax error.

\paragraph{(V2) KB Inconsistency:} A logical form $l$ may be inconsistent with the schema of $G$.
We identify semantic errors of different categories, such as type-incompatibility and schema hallucinations, implemented using rules over $l$ and $G$. 
These are also strong verifiers. 

\paragraph{(V3) Question-Logical Form Disagreement: }
This verifier checks if $l$ is semantically equivalent to the original natural language question $q$.
In Fig.~\ref{fig:FUn}, LF3 for KB1 disagrees with $q$.
This is a weak verifier.
First, $q$ may not have any equivalent logical form for $G$ due to intrinsic ambiguities even when it is answerable. 
For example, $q$ mentions a \textsc{person} {\em from a} \textsc{country}, when $G$ has the relations {\em born in} and {\em works in} between these types.
Secondly, this verifier is a probabilistic classifier that naturally makes occasional mistakes.
We define equivalence check between $l$ and $q$ using a multi-stage LLM pipeline, involving {\bf naturalization} of $l$ to $l^n$, {\bf back-translation} of $l^n$ to natural language question $q^b$ and {\bf semantic equivalence check} between $q$ and $q^b$.
More details are in the Appendix (Sec.\ref{subsec_appendix_error_check}).

\paragraph{(V4) Answer Inconsistency: } 
This verifier executes $l$ over $G$ to obtain an answer $A$ and then checks its compatibility with $q$.
This may fail for different reasons, such as 
{\bf (V4a)} $A$ containing an entity mentioned in $q$,
{\bf (V4b)} $A$ being empty, and others. 
Note that V4a is a strong verifier while V4b is weak, since an empty answer is valid for unanswerable questions (as for LF3 for KB2), but invalid for answerable ones.

\paragraph{Identifying Candidate Logical Forms: } 
Unless some logical form passes all checks and is therefore returned (Algo.~\ref{algo:fun} line 15), FUn constructs a candidate set $L$ of logical forms that are potentially flawed but not certainly so.
For our specific suite of weak verifiers, $l^{(i)}$ is added to $L$ if it passes one of 
V4b ($A$ is non-empty) as for LF3 for KB1, or V3 ($l^{i}$ is equivalent to $q$), as for LF3 for KB2.

\paragraph{Self Consistency for Unanswerability (scUn): }
Given a candidate set $L$ of logical forms and a question $q$, scUn assesses if the best candidate $l^*\in L$ has sufficient confidence.
If so, it outputs ($l=l^*$, $A=A^*$), as for KB2, $A^*$ being the answer from executing $l^*$ (may be \texttt{NA}).
Otherwise, scUn outputs ($l=$ \texttt{NK}, $A=$ \texttt{NA}), as for KB1.
For identifying the consensus choice from $L$, one possibility is self-consistency (sc)~\cite{DBLP:conf/iclr/0002WSLCNCZ23,DBLP:journals/corr/abs-2311-17311} that considers the answer for each $l\in L$, and returns those with the most common answer.
This requires {\em some answer} to accumulate enough probability by aggregation over reasoning paths.
However, for unanswerable questions, no single answer accumulates sufficient probability, and sc returns some low probability answer.

To address this, scUn first identifies via execution the most popular {\em non-empty} answer $A^*$ among logical forms in $L$, and decides using a threshold $t$ if it has enough supporters in $L$ (we use $t=\lfloor \frac{|L|}{2} \rfloor$).
If so, scUn uses LLM prompting to select the most appropriate supporting logical form $l^*\in L$ considering $q$, and outputs ($l=l^*$, $A=A^*$).
However, for KB1, the 3 logical forms among the candidates have 3 different answers, and therefore no consensus emerges ($\lfloor \frac{|L|}{2} \rfloor=1$).
Here, scUn considers logical forms from $L$ that agree on $A=$ \texttt{NA}.
If there are multiple such candidates, scUn selects the most suitable candidate $l^*$, again using LLM prompting, and outputs ($l=l^*$, $A=$ \texttt{NA}).
If there is no such candidate, scUn outputs ($l=$ \texttt{NK}, $A=$ \texttt{NA}).
For KB2 in the example, scUn selects LF3 -- the only logical form with empty answer.
Further details on scUn are in the Appendix (Sec.~\ref{subsec:scun_algo}).


\section{Experiments} \label{sec:experiments}

\begin{table*}[t]
\begin{center}
\small
\resizebox{0.99\textwidth}{!}{%
\begin{tabular}{l|ccccccc|cccccccc}
\hline
& \multicolumn{7}{c|}{{\bf WebQSP $\rightarrow$ GrailQAbility}} & \multicolumn{7}{c}{{\bf WebQSP $\rightarrow$ GraphQAbility}} \\
\multicolumn{1}{l|}{{\bf Model}} & \multicolumn{2}{c}{{\bf Overall}} & \multicolumn{2}{c}{{\bf Answerable}} & \multicolumn{3}{c|}{{\bf Unanswerable}} & \multicolumn{2}{c}{{\bf Overall}} & \multicolumn{2}{c}{{\bf Answerable}} & \multicolumn{3}{c}{{\bf Unanswerable}} \\
& {\bf F1} & {\bf EM-s} & {\bf F1} & {\bf EM-s} & {\bf F1(L)} & {\bf F1(R)} & {\bf EM-s} & {\bf F1} & {\bf EM-s} & {\bf F1} & {\bf EM-s} & {\bf F1(L)} & {\bf F1(R)} & {\bf EM-s}  \\
\hline
RetinaQA & 58.4 & 42.2 & 28.7 & 26.0 & \textbf{88.0} & \textbf{84.8} & 58.4 & 49.7 & 35.8 & 18.7 & 15.2 & 80.7 & 78.7 & 56.4 \\
Pangu & 54.5 & 43.8 & 31.2 & 29.6 & 83.8 & 80.4 & 58.0 & 53.4 & 33.0 & 30.3 & 26.4 & 76.5 & 74.8 & 39.6 \\
\hline
FuSIC-KBQA-U & \textbf{76.6} & 48.2 & {\bf 67.5} & 59.2 & 85.6 & 80.4 & 37.2 & 67.5 & 34.8 & 49.3 & 40.0 & 85.7 & 82.8 &  29.6\\
KB-Binder & 43.7 & 33.0 & 19.5 & 16.5 & 67.9 & 66.5 & 49.5 & 44.3 & 36.1 & 27.5 & 21.6 & 61.0 & 61.0 &  50.7 \\
\hline
\sys{} & {\bf 76.6} & {\bf 60.2} & 67.1 & {\bf 61.2} & 85.1 & 80.0 & \textbf{59.2} & \textbf{70.0} & \textbf{53.8} & {\bf 50.7} & {\bf 42.8} & {\bf 89.2} & \textbf{86.5} & \textbf{64.8} \\
\hline
\end{tabular}
}
\caption{Performance of different models on two datasets for few-shot KBQA transfer with unanswerability. 
{\bf Answerable} and {\bf Unanswerable} record performance for corresponding subsets and {\bf Overall} for the entire dataset.
}
\label{tab:unanswerability_main}
\end{center}
\end{table*}

We now present experimental evaluation of \sys{}. 
First, for few-shot KBQA transfer {\em with unanswerability}, we address the following research questions.
{\bf (R1)} How does \sys{} compare against SoTA KBQA models suitably adapted for this setting?
{\bf (R2)} How does \sys{} perform across different categories of unanswerability?
{\bf (R3)} How do the different components of \sys{} contribute to its performance?
Then, for {\em answerable} KBQA few-shot transfer, we ask:
{\bf (R4)} How does \sys{} compare against SoTA KBQA models for this setting?

\subsection{Experimental Setup}

\paragraph{Datasets:}
For in-domain and answerable KBQA, the three most popular datasets are GrailQA~\cite{gu:www2021grailqa}, GraphQA~\cite{su:emnlp2016-graphqa} and WebQSP~\cite{yih:acl2016webqsp}.
All of these have the same back-end KB (Freebase). 
For few-shot KBQA transfer, the only available datasets also have only answerable questions~\cite{patidar-etal-2024-fusic}.
GrailQAbility is the only available KBQA dataset with unanswerable questions~\cite{patidar-etal-2023-knowledge}.
This was constructed starting from GrailQA~\cite{gu:www2021grailqa} by systematically deleting schema and data elements from the back-end KB to introduce different categories of unanswerability into the queries. 

Our task needs source-target pairs, where the target contains unanswerable questions as well.
We construct our own transfer datasets using existing ones. 
For the transfer task to be non-trivial, the various distributions in the source and target need to be sufficiently dissimilar. 
WebQSP contains real user questions, which are manually annotated with logical forms, unlike GraphQA and GrailQA in which algorithmically generated logical forms are verbalized by crowd-workers. 
Since the source needs only answerable questions, we use WebQSP as source.
We select {\bf GrailQAbility} as one of our targets, since it already contains unanswerable questions, to create the {\bf WebQSP$\rightarrow$GrailQAbility} dataset.
We create our second target dataset using GraphQA, by introducing unanswerability into it. 
We do so by replacing its KB with the modified KB in GrailQAbility, which renders a subset of questions unanswerable. 
We label these appropriately as schema-level or data-level unanswerable.
We name this dataset {\bf GraphQAbility}.
Using this, we create the {\bf WebQSP$\rightarrow$GraphQAbility} dataset.
The WebQSP training set has 2,858 labeled questions.
We create the test sets for GrailQAbility and GraphQAblity by selecting 250 answerable and 250 unanswerable questions uniformly at random from the GrailQAbility and GraphQA test sets. 
We create few-shots by selecting 100 questions (50 answerable and 50 unanswerable) uniformly at random from the GrailQAbility dev set and GraphQA train set respectively. 

The test sets of both datasets have 50\% each of answerable and unanswerable questions. 
Of the unanswerable questions, the percentages of schema-level and data-level unanswerable are 66\% and 34\% in WebQSP$\rightarrow$GrailQAbility and 51.6\% and 48.4\% in WebQSP$\rightarrow$GraphQAbility. 
Additionally, the average number of relations per logical form is higher for GraphQAbility than for GrailQAbility, while it is the reverse for questions (using average number of tokens).
This suggests that GraphQAbility is harder for few-shot transfer, requiring more reasoning with shorter context.
Other statistics for the datasets are in Tab.~ \ref{tab:dataset_comparision} and discussed in the Appendix (Sec.~\ref{appendix:dataset}).

\begin{table*}[t]
\centering
\small
\begin{tabular}{l|cccccc|cccccc}
\hline
& \multicolumn{6}{c|}{{\bf WebQSP $\rightarrow$ GrailQAbility}}& \multicolumn{6}{c}{{\bf WebQSP $\rightarrow$ GraphQAbility}}\\

{\bf Model}&  \multicolumn{3}{c}{{\bf Schema Level}}&  \multicolumn{3}{c|}{{\bf Data Level}}&  \multicolumn{3}{c}{{\bf Schema Level}}& \multicolumn{3}{c}{{\bf Data Level}}\\ 
& {\bf F1(L)} & {\bf F1(R)} & {\bf EM-s} & {\bf F1(L)} & {\bf F1(R)} & {\bf EM-s} & {\bf F1(L)} & {\bf F1(R)} & {\bf EM-s}& {\bf F1(L)} & {\bf F1(R)} &{\bf EM-s}\\  
\hline 
RetinaQA& \textbf{94.1}& \textbf{90.9}& \textbf{79.4}& 76.3& 72.9& 14.1& 83.2& 82.0& 72.3& 73.7& 72.7&12.1\\
Pangu& 91.1 & 87.9 & 87.9 & 69.6 & 65.9 & 00.0 & 77.3 & 74.4 & 74.4 & 73.3 & 72.7 & 00.0\\
\hline
FuSIC-U& 85.4& 80.6& 30.9& \textbf{86.0}& \textbf{80.0}& \textbf{49.4}& 86.6& 82.6& 19.0& \textbf{83.3}& \textbf{83.3}&\textbf{51.5}\\
KB-Binder & 75.1 & 73.9 & 70.1 & 53.1 & 51.5 & 09.5 & 67.0 & 65.9 & 60.9 & 41.2 & 41.2 & 06.8 \\
\hline
\sys{}& 85.8& 81.2& 70.9& 83.8& 77.6& 36.5& \textbf{92.4}& \textbf{87.5}& \textbf{75.6}& 80.3& 80.3&34.8\\
\hline
\end{tabular}
\caption{Model performance for categories of unanswerable questions. FuSIC-U is short hand for FuSIC-KBQA-U.}
\label{tab:unanswerability_categories}
\end{table*}

\paragraph{Models for comparison:}
As few-shot transfer for KBQA with unanswerability is a novel task, there are no existing baselines.
For {\em in-domain KBQA with unanswerability}, 
{\bf RetinaQA}~\cite{faldu:acl2024} and the unanswerability-adapted version of {\bf Pangu}~\cite{gu-etal-2023-dont} are the SoTA models.
For these, we use the available code.\footnote{\url{https://github.com/dair-iitd/RetinaQA}}$^{,}$\footnote{\url{https://github.com/dki-lab/Pangu}}
More details are in the Appendix (Sec.~\ref{inference_details}).

{\bf \sysold{}} is the SoTA model for few-transfer for KBQA with only answerable questions.
{\bf KB-Binder}~\cite{li:acl2023kbbinder}  is the SoTA for in-domain few-shot KBQA. 
Overall, \sysold{} and KB-Binder outperform all other supervised and LLM-equipped KBQA models adapted for few-shot transfer~\cite{patidar-etal-2024-fusic}.
We use available code for KB-Binder\footnote{\url{https://github.com/ltl3A87/KB-BINDER}}, and our own implementation for \sysold{}.
To adapt these two baselines for unanswerability, for fair comparison, we modify their logical form generation prompt in the same fashion as PUn for \sys{}.
Additionally, for \sysold{}, we remove execution-guided feedback (EGF) since it fails for unanswerability.
We denote this model {\bf \sysold{}-U}.
Observe that \sysold{}-U can also be seen as an ablation of \sys{}, without FUn.
More details about KB-Binder and \sysold{} are in the Appendix (Sec.~\ref{subsec:appendix_kbbinder}).

We use $\mathcal{L}=$\texttt{gpt-4-0613} for all LLM-equipped models.
For fair comparison, we allocate to all such models the same maximum aggregated prompt length for a question.
This is satisfied by equipping \sys{} with zero-shot generation and $n=4$ FUn iterations, \sysold{}-U with 5-shot generation and KB-Binder with 25-shot generation.

Though \sys{} and \sysold{ } allow flexible use of multiple supervised retrievers, for meaningful comparison with RetinaQA, we adapt RetinaQA as retriever for \sys{} and \sysold{}-U.
More details about \sysold{}'s retriever and compute infrastructure are in the Appendix (Sec.~\ref{comp_and_infra}).

\paragraph{Evaluation Measures:}

For KBQA as semantic parsing task, evaluation of logical forms is primary. 
For this, the existing EM measure~\cite{ye:acl2022rngkbqa} is defined only for logical forms represented using s-expressions. 
\sysold{}-U and \sys{} output logical forms in SPARQL, and Pangu, RetinaQA and KB-Binder in s-expression. 
So we propose a new measure {\bf EM-s} that checks {\em approximate equivalence} for a pair of programs either in SPARQL or s-expression.
More details are in the Appendix (Sec.~\ref{em_for_sparql}).

As in standard KBQA evaluation, we also evaluate answers. 
This is a {\em secondary evaluation} for giving the benefit of the doubt for getting the right answer, possibly via a logical form not equivalent to the gold-standard according to EM-s.
For answer evaluation in KBQA with unanswerability, ~\cite{patidar-etal-2023-knowledge} introduced lenient F1, denoted F1(L), in addition to regular F1, denoted F1(R). 
F1(L) relaxes F1(R) by not penalize the original answer for the complete KB.
Note that obtaining the right answer by chance has much higher probability than for logical forms, particularly for unanswerable questions with \texttt{NA} as the correct answer. 

\begin{table*}[ht]
\centering
\resizebox{2.1\columnwidth}{!}{\begin{tabular}{l|cccccccc|cccccccc}
\hline
& \multicolumn{8}{c|}{{\bf WebQSP $\rightarrow$ GrailQAbility}}& \multicolumn{8}{c}{{\bf WebQSP $\rightarrow$ GraphQAbility}}\\

{\bf Model}&  \multicolumn{2}{c}{{\bf Answerable}} &  \multicolumn{3}{c}{{\bf Schema L. UnAnswerable}}&  \multicolumn{3}{c|}{{\bf Data L. UnAnswerable}} 
&  \multicolumn{2}{c}{{\bf Answerable}} &  \multicolumn{3}{c}{{\bf Schema Level UnAns}}& \multicolumn{3}{c}{{\bf Data Level UnAns}}\\ 

& {\bf F1} & {\bf EM-s} & {\bf F1(L)} & {\bf F1(R)} & {\bf EM-s} & {\bf F1(L)} & {\bf F1(R)} & {\bf EM-s} 
& {\bf F1(L)} & {\bf EM-s} &{\bf F1(L)} & {\bf F1(R)} & {\bf EM-s}& {\bf F1(L)} & {\bf F1(R)} &{\bf EM-s}\\  
\hline 
\sys{}                  & 74.0 & 70.0 & 90.9 & 87.9 & 75.8 & 64.7 & 64.7 & 11.8 
                        & 59.0 & 48.0 & 97.1 & 91.4 & 80.0 & 86.7 & 86.7 & 26.3\\
 scUn $\Rightarrow$ sc  & 74.0 & 70.0 & 73.0 & 72.7 & 33.3 & 58.8 & 52.9 & 23.5 
                        & 64.0 & 54.0 & 74.4 & 68.6 & 22.9 & 80.0 & 80.0 & 13.3\\
                        
w/o syntax              & 67.3& 64.0 & 87.9& 90.9& 42.4& 76.5& 70.6& 17.7
                        & 57.0 & 46.0 & 81.0& 76.9& 26.9& 75.8& 75.0 & 16.7\\
                        
w/o kb-inc              & 70.3& 66.0 & 90.9& 90.9& \multicolumn{1}{c}{9.1} & 76.5 & 70.6& 29.4
                        & 51.7 & 42.0 & 100.0& 100.0& \multicolumn{1}{c}{4.2} & 79.5 & 75.0 & 20.8\\
                        
w/o q-lf                & 71.0 & 68.0 & 69.7 & 63.6& \multicolumn{1}{c}{0.0}  & 41.2 & 35.3& 5.9
                        & 47.7 & 38.0 & 69.5& 65.4& 07.7 & 67.0& 62.5 & 20.8\\

w/o ans-inc             & 72.0 & 70.0 & 85.9& 84.9& 33.3& 70.6& 70.6& 23.5
                        & 55.0 & 44.0 & 80.8& 76.9& 26.9& 79.5& 79.2& 25.0\\
\hline
\end{tabular}}
\caption{
Ablation performance of \sys{} (removing individual components with replacement) on subset of WebQSP $\rightarrow$ GraphQAbility. 
scUn $\Rightarrow$ sc denotes replacing scUn with self consistency. Other rows remove verifiers for syntax error (w/o syntax) (V1), KB inconsistency (w/o kb-inc) (V2), question logical form disagreement (w/o q-lf) (V3) and answer incompatibility (w/o ans-inc) (V4).
Evaluations are on 100 instances from test sets (50 answerable and 50 unanswerable questions sampled uniformly at random)}
\label{tab:unanswerability_ablation_full}
\end{table*}

\subsection{Unanswerability Setting}\label{subsec:exp_unanswerable}

We first address research question {\bf R1}.
Performances of different models for few-shot transfer with unanswerability are recorded in Tab.~\ref{tab:unanswerability_main}.
First, we observe that \sys{} significantly outperforms all baselines in terms of EM-s, and performs at par with \sysold{}-U and significantly better than all other models in F1. 
The other LLM-equipped models are not significantly better than the supervised models.
All the baselines perform almost at par for GraphQAbility, and KB-Binder performs worse than the other 3 for GrailQAbility.
This establishes usefulness of FUn equipped with scUn for few-shot transfer KBQA with unanswerability, beyond LLM-usage.
Secondly, each model trades off performance differently between answerable and unanswerable questions.
RetinaQA, Pangu and also KB-Binder fare better for unanswerable questions, while \sys{} and \sysold{}-U fare better for answerable ones.
However, \sys{} achieves the best balance across the two subsets.

We next briefly address research question {\bf R2}.
Performance of different models for different categories of unanswerability are recorded in Tab.~\ref{tab:unanswerability_categories}. 
All models struggle to fare well simultaneously for data-level and schema-level unanswerability.
\sysold{}-U performs the best for data-level while performing poorly (in terms of EM-s) for schema-level.
Conversely, RetinaQA performs well for schema-level, but has poor data-level EM-s.
\sys{} outperforms other models in schema-level unanswerability while being slightly worse in data-level unanswerability.
But among all models, it achieves the best tradeoff by far across unanswerability categories.

We next address research question {\bf R3}.
Tab.~\ref{tab:unanswerability_ablation_full} records the ablation analysis of \sys{}.
We see that the KB-Inconsistency verifier (V2) and the Q-LF Disagreement verifier (V3) lead to significant improvements in EM-s. 
The biggest benefit comes from V3 for both answerable and unanswerable questions.
Without V3, the correct LF for schema-level unanswerability is almost never generated, though answer accuracy stays high, indicating inability to reason.
Similarly, removing V2 reduces EM-s for schema-level unanswerable questions, with answer accuracy remaining high, indicating that answers are often correct despite flawed logical forms. 
The Answer Incompatibility verifiers (V4) also makes significant contributions to the performance. 
This analysis highlights the necessity of a mix of weak and strong verifiers for structural and semantic validity. 
Beyond verifiers, replacing scUn with self-consistency, as expected, leads to a drastic drop in unanswerable performance (though this comes with a benefit for answerable questions).

\begin{table}[h]
\begin{center}
    \small
    \centering
    \begin{tabular}{l|cc}
        
        \hline
        & {\bf WebQSP} $\rightarrow$  & {\bf WebQSP}  $\rightarrow$\\ 
        {\bf Model}  & {\bf GrailQA-Tech} & {\bf GraphQA-Pop} \\
        \hline
        \sysold{} & 70.8 & 52.3 \\ 
        \sys{}(sc) & \textbf{73.6} & \textbf{67.0 }\\ 
        \hline
        \sysold{}-U & 62.6 & 43.4 \\
        \sys{}(scUn) & \textbf{71.2}& \textbf{65.0}\\
        \hline
    \end{tabular}
    \caption{Performance using F1 of different models for few-shot KBQA transfer with only answerable questions.
    The models in the top block have prior knowledge of answerability, while those in the bottom block do not.}
    \label{tab:answerable}
\end{center}
\end{table}

Finally, we report accuracy for the weak verifiers. 
The Q-LF Disagreement Verifier (V3) has accuracies of 90\% and 88\% overall for WebQSP$\rightarrow$GraphQAbility and WebQSP$\rightarrow$GrailQAbility, with its back-translation component has 90\% and 94\%.
The accuracy of the Empty Answer Verifier (V4b) depends on the nature and fraction of unanswerable questions. 
Its accuracies are 75\% and 68\% for the two datasets, corresponding to  $\sim$25\% and $\sim$17\%of questions respectively with $l^*\neq$ \texttt{NK} and $A^*=$ \texttt{NA}.  
More details are in the Appendix (Sec.~\ref{appendix:subsec:weak-verifiers}).

\subsection{Answerable Setting}

We now address research question {\bf R4} for answerable-only KBQA transfer.
We use two datasets from existing literature~\cite{patidar-etal-2024-fusic}, including the hardest one (WebQSP $\rightarrow$ GraphQA-Pop).\footnote{\url{https://github.com/dair-iitd/FuSIC-KBQA/}}
For enabling comparison with earlier results, we use TIARA~\cite{shu-etal-2022-tiara} as the retriever for all models in this experiment.

This setting admits two sub-cases:
{\bf (A)} the models have knowledge that all questions are answerable, and {\bf (B)} though all questions are answerable, the models do not have this knowledge.

Setting (A) has been studied for KBQA~\cite{patidar-etal-2024-fusic}, and \sysold{} is the established SoTA model, outperforming a host of supervised and LLM-based models adapted for the task. 
To adapt for this setting, \sys{} requires three simplifications. (i) PUn is replaced with prompt for answerability, 
(ii) In FUn, V4b (empty answer) is moved from the set of weak verifiers to that of strong verifier, 
and (iii) scUn is replaced by standard self-consistency.

The first two rows in Tab.~\ref{tab:answerable} record performance for setting (A).
\sys{} significantly outperforms \sysold{} on both datasets, creating a new SoTA for this setting.
This shows the usefulness of iterative repair with a suite of strong and weak verifiers followed by self-consistency for few-shot KBQA transfer, even without unanswerability.

In the more realistic setting (B), which has not been studied before, the models make predictions assuming unanswerability. 
Here, we evaluate \sysold{}-U and \sys{} as in Sec.~\ref{subsec:exp_unanswerable}, only there are no truly unanswerable questions.
The bottom two rows of Tab.~\ref{tab:answerable} record the performance of the two models in this setting.
We see that \sys{} outperforms \sysold{} by a very large margin.
This further establishes the usefulness of scUn when guarantees about answerability are not available.

\subsection{Error Analysis}

\begin{table}
    \centering
    \small
    \begin{tabular}{lll|r}
    \hline
     EM-s $<1$    &  &  & 46.2\\
         &Retr. Err.  &  & 23.4\\
         &Gen. Err.  &  & 22.8\\
         &  &$l^*=$ \texttt{NK}, ${\hat l\neq }$ \texttt{NK}  & 8.4\\
         &  &$l^*\neq $ \texttt{NK}, ${\hat l=}$ \texttt{NK}  & 4.6\\
         &  &$l^*\neq$ \texttt{NK}, ${\hat l}\neq $\texttt{NK}, $l^*\neq {\hat l}$  & 9.8\\
         &  &  & \\
    \hline
    \end{tabular}
    \caption{FUn error analysis on WebQSP $\rightarrow$ GraphQAbility. $l^*$ \& ${\hat l}$ denote gold \& generated logical forms. Retrieval error means retrieval $r$ is missing $\geq 1$ KB elements (class, relation, entity) necessary for $l^*$. Generation error implies ${\hat l}\neq l^*$ despite correct retrieval. 
}
    \label{tab:error_analysis}
\end{table}

For WebQSP $\rightarrow$ GraphQAbility, we analyzed questions for which logical forms generated by \sys{} are incorrect (EM-s $<1$). 
Results are in Tab.~\ref{tab:error_analysis}.
We found three main causes for generation errors.
{\bf (1)} Some questions are inherently ambiguous, admitting multiple valid logical forms $l_1$ and $l_2$ in the original complete KB, though only one is recognized as the gold ($l^*=l_1$).
Deletion to introduce unanswerability eliminates $l_1$, so that that $l^*=$ \texttt{NK}, and the prediction ${\hat l}=l_2$ is unfairly penalized. 
{\bf (2)} Here, $l^*=l_1$ and the prediction ${\hat l}=l_2$, such that $l_1\neq l_2$ but are semantically equivalent. $l_1$ and $l_2$ are incorrectly judged non-equivalent by EM-s.
{\bf (3)} Here, FUn is unable to generate $l^*$ or any semantic equivalent of it within its iteration limit.

\section{Conclusions}
\label{sec:conclusion}

For real-world robust and low-resource KBQA, we have proposed the novel task of few-shot transfer learning with unanswerability.
We have introduced a new notion (FUn) of iterative feedback guided repair for answerable as well as unanswerable questions.
FUn (i) uses feedback from a diverse suite a strong and weak verifiers -- including a novel back-translation based verifier -- to create a set of candidate logical forms, and (ii) assesses this candidate set to either to detect unanswerability (and its category) or identify the best logical form using self consistency adapted for unanswerability (scUn).
We propose \sys{} that replaces the existing the iterative strategy, that assumes answerability of questions, with FUn in the SoTA few-shot answerable-only KBQA transfer model (\sysold{}).
Using two newly created datasets for this novel task, we show that \sys{} significantly outperforms adaptations of \sysold{} and other SoTA models for this setting, and also for answerable few-shot transfer KBQA. 

Our error analysis suggests that performing well across categories of unanswerability for few-shot transfer is still a challenge for KBQA and should be a focus of further research.
We have made our datasets and other resources public \footnote{\url{https://github.com/dair-iitd/FUn-FuSIC}}. 

\section*{Limitations}\label{sec:limitations} 

Since LLM inference involves randomness, experiments should ideally be repeated for multiple runs and results should report averages and error bars.
Unfortunately, we were not able to do this due to the prohibitive cost of GPT-4, and our results are based on single runs.

While GPT-4 is currently the best performing LLM, it is proprietary as well as expensive.
Ideally, evaluation should include open-source freely accessible LLMs as well.
We expect performance of all LLM-based approaches to drop when GPT-4 is replaced by a less powerful, open LLM.
Nonetheless, earlier research has shown that models with Mistral instead of GPT-4 still outperform fully supervised models for answerable few-shot transfer~\cite{patidar-etal-2024-fusic}.
Whether this trend holds for the unanswerable setting is an open question.
That said, following current trends, we expect the ability of open LLMs to steadily improve in the coming years.

\section*{Risks}\label{sec:risks} 

At the highest level, our work reduces risk compared to existing KBQA systems, which when inadequately adapted in a low-resource setting, incorrectly answer unanswerable questions, without acknowledging lack of knowledge.
However, can incorrectly inferring unanswerability, citing lack of knowledge when knowledge is in fact available, be a new type of risk?
While we cannot imagine such a risk at the present time, this may require more careful consideration.
In any case, KBQA models for unanswerability should strive to minimize this type of error, along with the other types.

\section*{Acknowledgments}

Mausam is supported by a contract with TCS, grants from IBM, Verisk, Huawei, Wipro, and the Jai Gupta chair fellowship by IIT Delhi. 
Indrajit would like to thank KnowDis AI for supporting his participation in the conference.
Riya would like to thank Graviton Research Capital for supporting her participation in the conference. 
The authors would like to thank the IIT-D HPC facility for its computational resources. 
We also thank Microsoft Accelerate Foundation Models Research (AFMR) program that provided us access to OpenAI models. 
We are also thankful to Mayur Patidar for helpful discussions.

\bibliography{main}

\begin{thebibliography}{33}
\expandafter\ifx\csname natexlab\endcsname\relax\def\natexlab#1{#1}\fi

\bibitem[{Cao et~al.(2022)Cao, Shi, Yao, Lv, Yu, Hou, Li, Liu, and Xiao}]{cao:acl2022progxfer}
Shulin Cao, Jiaxin Shi, Zijun Yao, Xin Lv, Jifan Yu, Lei Hou, Juanzi Li, Zhiyuan Liu, and Jinghui Xiao. 2022.
\newblock Program transfer for answering complex questions over knowledge bases.
\newblock In \emph{Proceedings of the 60th Annual Meeting of the Association for Computational Linguistics (Volume 1: Long Papers)}.

\bibitem[{Chen et~al.(2021)Chen, Liu, Yu, Lin, Lou, and Jiang}]{chen:acl2021retrack}
Shuang Chen, Qian Liu, Zhiwei Yu, Chin-Yew Lin, Jian-Guang Lou, and Feng Jiang. 2021.
\newblock {R}e{T}ra{C}k: A flexible and efficient framework for knowledge base question answering.
\newblock In \emph{Proceedings of the 59th Annual Meeting of the Association for Computational Linguistics and the 11th International Joint Conference on Natural Language Processing: System Demonstrations}.

\bibitem[{Chen et~al.(2023{\natexlab{a}})Chen, Aksitov, Alon, Ren, Xiao, Yin, Prakash, Sutton, Wang, and Zhou}]{DBLP:journals/corr/abs-2311-17311}
Xinyun Chen, Renat Aksitov, Uri Alon, Jie Ren, Kefan Xiao, Pengcheng Yin, Sushant Prakash, Charles Sutton, Xuezhi Wang, and Denny Zhou. 2023{\natexlab{a}}.
\newblock \href {https://doi.org/10.48550/ARXIV.2311.17311} {Universal self-consistency for large language model generation}.
\newblock \emph{CoRR}, abs/2311.17311.

\bibitem[{Chen et~al.(2023{\natexlab{b}})Chen, Lin, Schaerli, and Zhou}]{chen2023teaching}
Xinyun Chen, Maxwell Lin, Nathanael Schaerli, and Denny Zhou. 2023{\natexlab{b}}.
\newblock Teaching large language models to self-debug.
\newblock In \emph{The 61st Annual Meeting Of The Association For Computational Linguistics}.

\bibitem[{Das et~al.(2022)Das, Godbole, Naik, Tower, Zaheer, Hajishirzi, Jia, and Mccallum}]{das:icml2022subgraphcbr}
Rajarshi Das, Ameya Godbole, Ankita Naik, Elliot Tower, Manzil Zaheer, Hannaneh Hajishirzi, Robin Jia, and Andrew Mccallum. 2022.
\newblock Knowledge base question answering by case-based reasoning over subgraphs.
\newblock In \emph{Proceedings of the 39th International Conference on Machine Learning}.

\bibitem[{Das et~al.(2021)Das, Zaheer, Thai, Godbole, Perez, Lee, Tan, Polymenakos, and McCallum}]{das:emnlp2021cbr}
Rajarshi Das, Manzil Zaheer, Dung Thai, Ameya Godbole, Ethan Perez, Jay~Yoon Lee, Lizhen Tan, Lazaros Polymenakos, and Andrew McCallum. 2021.
\newblock Case-based reasoning for natural language queries over knowledge bases.
\newblock In \emph{Proceedings of the 2021 Conference on Empirical Methods in Natural Language Processing}.

\bibitem[{Dong et~al.(2023)Dong, Zhang, Ge, Mao, Gao, lu~Chen, Lin, and Lou}]{dong2023c3zeroshottexttosqlchatgpt}
Xuemei Dong, Chao Zhang, Yuhang Ge, Yuren Mao, Yunjun Gao, lu~Chen, Jinshu Lin, and Dongfang Lou. 2023.
\newblock \href {http://arxiv.org/abs/2307.07306} {C3: Zero-shot text-to-sql with chatgpt}.

\bibitem[{Faldu et~al.(2024)Faldu, Bhattacharya, and Mausam}]{faldu:acl2024}
Prayushi Faldu, Indrajit Bhattacharya, and Mausam. 2024.
\newblock {RETINAQA} : A knowledge base question answering model robust to both answerable and unanswerable questions.
\newblock In \emph{Proceedings of the 62nd Annual Meeting of the Association for Computational Linguistics (Volume 1: Long Papers)}, Bangkok, Thailand. Association for Computational Linguistics.

\bibitem[{Grattafiori et~al.(2024)Grattafiori, Dubey, Jauhri, Pandey, Kadian, Al-Dahle, Letman, Mathur, Schelten, Vaughan, Yang, Fan, Goyal, Hartshorn, Yang, Mitra, Sravankumar, Korenev, Hinsvark, Rao, Zhang, and Rodriguez}]{grattafiori2024llama3herdmodels}
Aaron Grattafiori, Abhimanyu Dubey, Abhinav Jauhri, Abhinav Pandey, Abhishek Kadian, Ahmad Al-Dahle, Aiesha Letman, Akhil Mathur, Alan Schelten, Alex Vaughan, Amy Yang, Angela Fan, Anirudh Goyal, Anthony Hartshorn, Aobo Yang, Archi Mitra, Archie Sravankumar, Artem Korenev, Arthur Hinsvark, Arun Rao, Aston Zhang, and Aurelien Rodriguez. 2024.
\newblock \href {http://arxiv.org/abs/2407.21783} {The llama 3 herd of models}.

\bibitem[{Gu et~al.(2023)Gu, Deng, and Su}]{gu-etal-2023-dont}
Yu~Gu, Xiang Deng, and Yu~Su. 2023.
\newblock \href {https://doi.org/10.18653/v1/2023.acl-long.270} {Don{'}t generate, discriminate: A proposal for grounding language models to real-world environments}.
\newblock In \emph{Proceedings of the 61st Annual Meeting of the Association for Computational Linguistics (Volume 1: Long Papers)}, pages 4928--4949, Toronto, Canada. Association for Computational Linguistics.

\bibitem[{Gu et~al.(2021)Gu, Kase, Vanni, Sadler, Liang, Yan, and Su}]{gu:www2021grailqa}
Yu~Gu, Sue Kase, Michelle Vanni, Brian Sadler, Percy Liang, Xifeng Yan, and Yu~Su. 2021.
\newblock Beyond i.i.d.: Three levels of generalization for question answering on knowledge bases.
\newblock In \emph{Proceedings of the Web Conference 2021}, WWW '21.

\bibitem[{Li et~al.(2023)Li, Ma, Zhuang, Gu, Su, and Chen}]{li:acl2023kbbinder}
Tianle Li, Xueguang Ma, Alex Zhuang, Yu~Gu, Yu~Su, and Wenhu Chen. 2023.
\newblock \href {https://doi.org/10.18653/v1/2023.acl-long.385} {Few-shot in-context learning on knowledge base question answering}.
\newblock In \emph{Proceedings of the 61st Annual Meeting of the Association for Computational Linguistics (Volume 1: Long Papers)}, pages 6966--6980, Toronto, Canada. Association for Computational Linguistics.

\bibitem[{Madaan et~al.(2023)Madaan, Tandon, Gupta, Hallinan, Gao, Wiegreffe, Alon, Dziri, Prabhumoye, Yang, Gupta, Majumder, Hermann, Welleck, Yazdanbakhsh, and Clark}]{madaan2023selfrefine}
Aman Madaan, Niket Tandon, Prakhar Gupta, Skyler Hallinan, Luyu Gao, Sarah Wiegreffe, Uri Alon, Nouha Dziri, Shrimai Prabhumoye, Yiming Yang, Shashank Gupta, Bodhisattwa~Prasad Majumder, Katherine Hermann, Sean Welleck, Amir Yazdanbakhsh, and Peter Clark. 2023.
\newblock \href {http://arxiv.org/abs/2303.17651} {Self-refine: Iterative refinement with self-feedback}.

\bibitem[{Mitra et~al.(2022)Mitra, Ramnani, and Sengupta}]{mitra:naacl2022cmhopkgqa}
Sayantan Mitra, Roshni Ramnani, and Shubhashis Sengupta. 2022.
\newblock Constraint-based multi-hop question answering with knowledge graph.
\newblock In \emph{Proceedings of the 2022 Conference of the North American Chapter of the Association for Computational Linguistics: Human Language Technologies: Industry Track}.

\bibitem[{Nie et~al.(2024)Nie, Zhang, Wang, and Liu}]{nie:arxiv2023kbcoder}
Zhijie Nie, Richong Zhang, Zhongyuan Wang, and Xudong Liu. 2024.
\newblock \href {https://doi.org/10.1609/aaai.v38i17.29848} {Code-style in-context learning for knowledge-based question answering}.
\newblock \emph{Proceedings of the AAAI Conference on Artificial Intelligence}, 38(17):18833--18841.

\bibitem[{Olausson et~al.(2024)Olausson, Inala, Wang, Gao, and Solar-Lezama}]{olausson:iclr2024}
Theo~X. Olausson, Jeevana~Priya Inala, Chenglong Wang, Jianfeng Gao, and Armando Solar-Lezama. 2024.
\newblock \href {https://openreview.net/forum?id=y0GJXRungR} {Is self-repair a silver bullet for code generation?}
\newblock In \emph{The Twelfth International Conference on Learning Representations}.

\bibitem[{Paszke et~al.(2019)Paszke, Gross, Massa, Lerer, Bradbury, Chanan, Killeen, Lin, Gimelshein, Antiga, Desmaison, Kopf, Yang, DeVito, Raison, Tejani, Chilamkurthy, Steiner, Fang, Bai, and Chintala}]{pytorch}
Adam Paszke, Sam Gross, Francisco Massa, Adam Lerer, James Bradbury, Gregory Chanan, Trevor Killeen, Zeming Lin, Natalia Gimelshein, Luca Antiga, Alban Desmaison, Andreas Kopf, Edward Yang, Zachary DeVito, Martin Raison, Alykhan Tejani, Sasank Chilamkurthy, Benoit Steiner, Lu~Fang, Junjie Bai, and Soumith Chintala. 2019.
\newblock \href {https://proceedings.neurips.cc/paper/2019/file/bdbca288fee7f92f2bfa9f7012727740-Paper.pdf} {Pytorch: An imperative style, high-performance deep learning library}.
\newblock In \emph{Advances in Neural Information Processing Systems}, volume~32. Curran Associates, Inc.

\bibitem[{Patidar et~al.(2023)Patidar, Faldu, Singh, Vig, Bhattacharya, and ~}]{patidar-etal-2023-knowledge}
Mayur Patidar, Prayushi Faldu, Avinash Singh, Lovekesh Vig, Indrajit Bhattacharya, and Mausam ~. 2023.
\newblock \href {https://doi.org/10.18653/v1/2023.acl-long.576} {Do {I} have the knowledge to answer? investigating answerability of knowledge base questions}.
\newblock In \emph{Proceedings of the 61st Annual Meeting of the Association for Computational Linguistics (Volume 1: Long Papers)}, pages 10341--10357, Toronto, Canada. Association for Computational Linguistics.

\bibitem[{Patidar et~al.(2024)Patidar, Sawhney, Singh, Chatterjee, Mausam, and Bhattacharya}]{patidar-etal-2024-fusic}
Mayur Patidar, Riya Sawhney, Avinash~Kumar Singh, Biswajit Chatterjee, Mausam, and Indrajit Bhattacharya. 2024.
\newblock Few-shot transfer learning for knowledge base question answering: Fusing supervised models with in-context learning.
\newblock In \emph{Proceedings of the 62nd Annual Meeting of the Association for Computational Linguistics (Volume 1: Long Papers)}, Bangkok, Thailand. Association for Computational Linguistics.

\bibitem[{Pourreza and Rafiei(2023)}]{pourreza2023dinsqldecomposedincontextlearning}
Mohammadreza Pourreza and Davood Rafiei. 2023.
\newblock \href {http://arxiv.org/abs/2304.11015} {Din-sql: Decomposed in-context learning of text-to-sql with self-correction}.

\bibitem[{Ravishankar et~al.(2022)Ravishankar, Thai, Abdelaziz, Mihindukulasooriya, Naseem, Kapanipathi, Rossiello, and Fokoue}]{ravishankar:emnlp2022}
Srinivas Ravishankar, Dung Thai, Ibrahim Abdelaziz, Nandana Mihindukulasooriya, Tahira Naseem, Pavan Kapanipathi, Gaetano Rossiello, and Achille Fokoue. 2022.
\newblock A two-stage approach towards generalization in knowledge base question answering.
\newblock In \emph{Findings of the Association for Computational Linguistics: EMNLP 2022}.

\bibitem[{Saxena et~al.(2022)Saxena, Kochsiek, and Gemulla}]{saxena:acl2022}
Apoorv Saxena, Adrian Kochsiek, and Rainer Gemulla. 2022.
\newblock Sequence-to-sequence knowledge graph completion and question answering.
\newblock In \emph{Proceedings of the 60th Annual Meeting of the Association for Computational Linguistics (Volume 1: Long Papers)}.

\bibitem[{Shinn et~al.(2023)Shinn, Cassano, Gopinath, Narasimhan, and Yao}]{shinn:neurips2023reflexion}
Noah Shinn, Federico Cassano, Ashwin Gopinath, Karthik~R Narasimhan, and Shunyu Yao. 2023.
\newblock \href {https://openreview.net/forum?id=vAElhFcKW6} {Reflexion: language agents with verbal reinforcement learning}.
\newblock In \emph{Thirty-seventh Conference on Neural Information Processing Systems}.

\bibitem[{Shu and Yu(2024)}]{shu:arxiv2023bottlenecks}
Yiheng Shu and Zhiwei Yu. 2024.
\newblock \href {https://aclanthology.org/2024.eacl-srw.7} {Distribution shifts are bottlenecks: Extensive evaluation for grounding language models to knowledge bases}.
\newblock In \emph{Proceedings of the 18th Conference of the European Chapter of the Association for Computational Linguistics: Student Research Workshop}, pages 71--88, St. Julian{'}s, Malta. Association for Computational Linguistics.

\bibitem[{Shu et~al.(2022)Shu, Yu, Li, Karlsson, Ma, Qu, and Lin}]{shu-etal-2022-tiara}
Yiheng Shu, Zhiwei Yu, Yuhan Li, B{\"o}rje Karlsson, Tingting Ma, Yuzhong Qu, and Chin-Yew Lin. 2022.
\newblock {TIARA}: Multi-grained retrieval for robust question answering over large knowledge base.
\newblock In \emph{Proceedings of the 2022 Conference on Empirical Methods in Natural Language Processing}.

\bibitem[{Slobodkin et~al.(2023)Slobodkin, Goldman, Caciularu, Dagan, and Ravfogel}]{slobodkin-etal-2023-curious}
Aviv Slobodkin, Omer Goldman, Avi Caciularu, Ido Dagan, and Shauli Ravfogel. 2023.
\newblock \href {https://aclanthology.org/2023.emnlp-main.220} {The curious case of hallucinatory (un)answerability: Finding truths in the hidden states of over-confident large language models}.
\newblock In \emph{Proceedings of the 2023 Conference on Empirical Methods in Natural Language Processing}, pages 3607--3625.

\bibitem[{Su et~al.(2016)Su, Sun, Sadler, Srivatsa, G{\"u}r, Yan, and Yan}]{su:emnlp2016-graphqa}
Yu~Su, Huan Sun, Brian Sadler, Mudhakar Srivatsa, Izzeddin G{\"u}r, Zenghui Yan, and Xifeng Yan. 2016.
\newblock \href {https://doi.org/10.18653/v1/D16-1054} {On generating characteristic-rich question sets for {QA} evaluation}.
\newblock In \emph{Proceedings of the 2016 Conference on Empirical Methods in Natural Language Processing}, pages 562--572, Austin, Texas. Association for Computational Linguistics.

\bibitem[{Wang et~al.(2023)Wang, Wei, Schuurmans, Le, Chi, Narang, Chowdhery, and Zhou}]{DBLP:conf/iclr/0002WSLCNCZ23}
Xuezhi Wang, Jason Wei, Dale Schuurmans, Quoc~V. Le, Ed~H. Chi, Sharan Narang, Aakanksha Chowdhery, and Denny Zhou. 2023.
\newblock \href {https://openreview.net/pdf?id=1PL1NIMMrw} {Self-consistency improves chain of thought reasoning in language models}.
\newblock In \emph{The Eleventh International Conference on Learning Representations, {ICLR} 2023, Kigali, Rwanda, May 1-5, 2023}. OpenReview.net.

\bibitem[{Wang et~al.(2022)Wang, Srinivasan, and Jin}]{wang:naacl2022mhopkgqa}
Yu~Wang, Vijay Srinivasan, and Hongxia Jin. 2022.
\newblock A new concept of knowledge based question answering ({KBQA}) system for multi-hop reasoning.
\newblock In \emph{Proceedings of the 2022 Conference of the North American Chapter of the Association for Computational Linguistics: Human Language Technologies}.

\bibitem[{Wolf et~al.(2020)Wolf, Debut, Sanh, Chaumond, Delangue, Moi, Cistac, Rault, Louf, Funtowicz, Davison, Shleifer, von Platen, Ma, Jernite, Plu, Xu, Scao, Gugger, Drame, Lhoest, and Rush}]{wolf-etal-2020-transformers}
Thomas Wolf, Lysandre Debut, Victor Sanh, Julien Chaumond, Clement Delangue, Anthony Moi, Pierric Cistac, Tim Rault, Rémi Louf, Morgan Funtowicz, Joe Davison, Sam Shleifer, Patrick von Platen, Clara Ma, Yacine Jernite, Julien Plu, Canwen Xu, Teven~Le Scao, Sylvain Gugger, Mariama Drame, Quentin Lhoest, and Alexander~M. Rush. 2020.
\newblock \href {https://www.aclweb.org/anthology/2020.emnlp-demos.6} {Transformers: State-of-the-art natural language processing}.
\newblock In \emph{Proceedings of the 2020 Conference on Empirical Methods in Natural Language Processing: System Demonstrations}, pages 38--45, Online. Association for Computational Linguistics.

\bibitem[{Ye et~al.(2022)Ye, Yavuz, Hashimoto, Zhou, and Xiong}]{ye:acl2022rngkbqa}
Xi~Ye, Semih Yavuz, Kazuma Hashimoto, Yingbo Zhou, and Caiming Xiong. 2022.
\newblock {RNG}-{KBQA}: Generation augmented iterative ranking for knowledge base question answering.
\newblock In \emph{Proceedings of the 60th Annual Meeting of the Association for Computational Linguistics (Volume 1: Long Papers)}.

\bibitem[{Yih et~al.(2016)Yih, Richardson, Meek, Chang, and Suh}]{yih:acl2016webqsp}
Wen-tau Yih, Matthew Richardson, Chris Meek, Ming-Wei Chang, and Jina Suh. 2016.
\newblock The value of semantic parse labeling for knowledge base question answering.
\newblock In \emph{Proceedings of the 54th Annual Meeting of the Association for Computational Linguistics (Volume 2: Short Papers)}.

\bibitem[{Zhang et~al.(2022)Zhang, Zhang, Yu, Tang, Tang, Li, and Chen}]{zhang:acl2022subgraph}
Jing Zhang, Xiaokang Zhang, Jifan Yu, Jian Tang, Jie Tang, Cuiping Li, and Hong Chen. 2022.
\newblock Subgraph retrieval enhanced model for multi-hop knowledge base question answering.
\newblock In \emph{Proceedings of the 60th Annual Meeting of the Association for Computational Linguistics (Volume 1: Long Papers)}.

\end{thebibliography}

\appendix
\section{Appendix}

\subsection{KBQA Elaboration}\label{subsec:appendix_problem_details}
Here we elaborate on different aspects of the KBQA problem.

\subsubsection{Challenges in Few Shot Transfer Learning for KBQA}
The source and target tasks may differ significantly.
First, the data and schema of the knowledge bases $G^t$ and $G^s$ and the domains they cover may be different. 
Secondly, the distributions of questions and logical forms defined over the KBs may be different in $D^t$ and $D^s$.

\subsubsection{Different Types of Unanswerability}\label{kbqa_unanswerability}
Unanswerable questions in KBQA can be  categorized into (a) Schema Level Unanswerability : the question does not have a corresponding logical form that is valid for the KB , (b) Data level unanswerability:  it has a valid logical form $l$ for the KB, but which on executing returns an empty answer. Schema level unanswerable questions can further be categorized into (1) Missing Class: The class/type required to construct the logical form is not defined for the KB, (2) Missing Relation: The relation required to construct the logical form is not defined for the KB, (3) Missing Topic Entity: The topic entity specified in the question is missing from the KB. 
Data level unanswerable questions can be categorized into (1) Missing entity: all classes and relations required to construct the logical form are present in the KB , but there exists no path from the topic entity node to the answer node in the KB due to missing intermediary entities (2) Missing Fact: all classes, relations and entities required to answer the question are present in the KB. However, the (subject, relation, object) path is not connected in the KB. 

\subsection{Algorithm for Self Consistency with Unanswerability}\label{subsec:scun_algo}

The high level algorithm for self consistency with Unanswerability (scUn) is described in Algo.~\ref{algo:scUn}.
\begin{algorithm}
\caption{ScUn$(q,L,\mathcal{L})$}
\label{algo:scUn}
\begin{algorithmic}[1] 
\State $(c, l, A) = \mbox{assessConf}(q, L, \mathcal{L})$
\State $\mbox{{\bf if} }(c)$, $\mbox{{\bf return}}(l, A)$
\State $\mbox{{\bf else} }$, $\mbox{{\bf return}}(\texttt{NK}, \texttt{NA})$
\end{algorithmic}
\end{algorithm}

The high level algorithm for assessing confidence in the set of candidate logical forms ($\mbox{assessConf}$) is described in Algo.~\ref{algo:scUn_assess_conf}.
We use abbreviations NE and E to indicate non-empty and empty respectively. popAnsNE is abbreviation of "(most) popular answer non-empty".

\begin{algorithm}
\caption{$\mbox{assessConf}(q,L,\mathcal{L})$}
\label{algo:scUn_assess_conf}
\begin{algorithmic}[1] 
\State $(c, L^p, A^p) = \mbox{popAnsNE}(L, t)$
\If{$(c)$}
    \State $l=\mbox{selectBestNE}(q,L^p,\mathcal{L})$
    \State $\mbox{{\bf return}}(\mbox{T}, l, A^p)$
\EndIf
\State $(c, L^p) = \mbox{popAnsE}(L, t)$
\If{$(c)$}
    \State $l=\mbox{selectBestE}(q,L^p,\mathcal{L})$
    \State $\mbox{{\bf return}}(\mbox{T}, l, \texttt{NA})$
\EndIf
\State $\mbox{{\bf return}}(\mbox{F}, \texttt{NK}, \texttt{NA})$
\end{algorithmic}
\end{algorithm}

\subsection{Details of FUn Verifiers}
\label{subsec_appendix_error_check}

Here we discuss the verifiers V2, V3 and V4 in more detail. 

\paragraph{(V2) Semantic Error (KB Inconsistency):} A syntactically correct logical form $l$ may still be inconsistent with the schema of $G$.
This is a likely error even for SoTA LLMs since these are unfamiliar with the specific KB $G$.
Semantic errors have different categories, such as type-incompatibility, schema hallucinations. 
{\bf (V2a)} Incompatibility in types: $l$ contains a variable and a connecting relation whose types are incompatible in $G$. 
This is the case for LF1 for all three KBs in the example.
{\bf (V2b)} Schema hallucinations: $l$ contains schema elements (types, relations, entities) absent in $G$.  
{\bf (V2c)} Type casting errors: Literals in $l$ are not correctly type cast for $G$, e.g. numeric literals as float for Freebase. 
All of these are certain checks, and are implemented using rules defined over $l$ and $G$.
The feedback mentions the type of error and the specifics, e.g., the hallucinated relation, or the incompatible type-relation pair.

\paragraph{(V3) Question-Logical form Disagreement: }
FUn performs equivalence check between $l$ and $q$ using a novel multi-stage LLM pipeline.
(i) The variable names in $l$ are first naturalized to $l^n$ considering $q$ and preserving semantics, e.g. by replacing {\em '?x'} with {\em '?actor'}.
(ii) $l^n$ is back-translated into a natural language question $q^b$.
(iii) $q^b$ is finally checked for semantic equivalence with $q$.
The first two steps are performed using zero-shot prompting, while the last is performed using few-shots constructed using the target few-shots $D^t$. 
The feedback mentions lack of equivalence as the type of error.

\paragraph{(V4) Answer Inconsistency: } If the $l$ is syntactically and semantically correct, it is executed over $G$ to obtain an answer $a$. 
$a$ is then checked for compatibility with $q$.
This may fail for different reasons.
{\bf (V4a)} $a$ (which is a set in general) contains an entity also in $l$ and therefore mentioned $q$, which is an aberration.
{\bf (V4b)} $a$ is empty, as in LF3 for KB2 in the example.
All of these checks are implemented using rules defined over $l$ and $G$.
Note that while the first two are certain checks, the last is not.
An empty answer is valid for unanswerable questions, as for LF3 for KB2, but invalid for answerable ones.
As before, the feedback mentions the type of error and the specifics.

\begin{table*}[h]
\centering
\small
\begin{tabular}{l|cc|cc|cc}
\hline
 & \multicolumn{2}{c|}{{Knowledge Base}} & \multicolumn{2}{c|}{{Logical Forms}}  & \multicolumn{2}{c}{{NL Questions}} \\

{\bf Source$\rightarrow$ Target}&  Domain JS & New Rel\% & Function JS & $\#$Relation  & Src-Tgt  & $\#$Token \\
&            &         &             & SrcAv/TgtAv & Cosine Sim & SrcAv/TgtAv\\
\hline
WebQSP$\rightarrow$ GrailQAbility & 0.67 & 93.8 & 0.61 & 1.69/1.45 & 0.34 & 6.60/11.27\\

WebQSP$\rightarrow$ GraphQAbility & 0.67 & 93.2 & 0.42 & 1.69/1.64 & 0.32 & 6.60/9.69 \\
\hline
\end{tabular}
\caption{ Statistics for different source and target (test set) KBQA task pairs in terms of the knowledge base, logical forms and natural language questions. ‘Domain(JS)’ is Jensen Shannon (JS) divergence between domain distribution of questions, ‘New Rel\%’ is percentage of questions in target with new (unseen) relations, ‘Function (JS)’ is JS-divergence between distributions over functions in logical forms, ‘$\#$Relations’ shows the source average and the target average for number of relations per logical form, ‘Src-Tgt Cosine Sim’ is average minimum cosine distance between source and target questions and ‘$\#$Tokens’ shows the source average and target average of number of tokens per question.}
\label{tab:dataset_comparision}
\end{table*}

\subsection{Additional Dataset Statistics}\label{appendix:dataset}

Here we include additional statistics on the two datasets WebQSP$\rightarrow$GrailQAbility and WebQSP$\rightarrow$GraphQAbility. 
We quantify different measures of hardness for the datasets. 
The results are tabulated in Tab.~\ref{tab:dataset_comparision}.
Tab.~\ref{tab:unanswerablility_distribution} shows percentage of the two categories of unanswerable questions.

\begin{table}[h]
\begin{center}
    \small
    \centering
    \begin{tabular}{c|cc}
        
        \hline
        {\bf Dataset} & Schema level  & Data level\\ 
         & UnAns & UnAns \\
        \hline
        WebQSP$\rightarrow$ GrailQAbility & 34.0 & 66.0 \\ 
        WebQSP$\rightarrow$ GraphQAbility & 48.4 & 51.6\\ 
        \hline
    \end{tabular}
    \caption{Percentages of schema and data level unanswerability among unanswerable questions.}
    \label{tab:unanswerablility_distribution}
\end{center}
\end{table}

\subsection{EM-s: Automated Approximate Equivalence Check for SPARQL} \label{em_for_sparql}
As has been observed in \cite{patidar-etal-2023-knowledge}, answer evaluation by itself is not a robust measure for evaluation of KBQA models when the dataset contains unanswerability. Traditional KBQA models that generate s-expressions can be evaluated using EM, which checks for logical form equivalence between two logical forms, since it is possible to compare equivalence between two s-expressions efficiently. However, \sys{} generates SPARQL queries instead. Directly comparing program equivalence between two SPARQL queries is an undecidable problem \footnote{\url{https://users.dcc.uchile.cl/~cgutierr/papers/expPowSPARQL.pdf}}. 
\citet{patidar-etal-2024-fusic} suggest a semi-automatic strategy for comparison of sparql queries. 
We propose a completely automatic metric for SPARQL equivalence check. 
Two SPARQL queries are equivalent by the EM-s check if (a) the relations occurring in the two queries are same. (b) the entities occurring in the two queries are the same (c) the answer set obtained by executing the queries over the KB are the same. Note that the EM-s check is necessary, but not sufficient for two SPARQL queries to be equivalent. 

Since these are a necessary but not sufficient condition for logical form equivalence, we compared EM-s with EM, where both are applicable and found $>98\%$ agreement.

\subsection{Model Evaluation: Additional Details}

In this section, we do a deeper evaluation of performance of different models across different categories of unanswerability, as explained in \cite{patidar-etal-2023-knowledge}. There are two broad categories of unanswerability --- schema level unanswerability (absence of knowledge in terms of KB ontology or entities required to construct the logical form) and data level unanswerability (absence of facts or intermediate entities of the logical  form path on the KB). 

We expect that (a) due to poor ability of supervised models to generalize in transfer learning settings, RetinaQA will struggle to generate correct logical forms for data level unanswerable questions, and (b) due to the strong generalization ability of \sysold{}, it should be able to perform well for data level unanswerable questions. 
However, since it is biased towards returning incorrect logical forms instead of abstaining from returning a logical form, it will perform poorly at identifying schema level unanswerable questions. (c) \sys{} should be able to maintain the performance of \sysold{} on data level unanswerable questions to a large extent, while significantly improving the performance on schema level unanswerable questions. 

Performance on the WebQSP $\rightarrow$GrailQAbility and WebQSP $\rightarrow$GraphQAbility datasets show that the trends are indeed as expected. 

\subsection{FUn Cost Analysis for Proprietary LLMs}

\sysold{}, as well as the adapted versions of \sysold{}, such as FuSIC-KBQA-U and \sys{} rerank the classes, relations and paths. 
The total cost for reranking for one question is \$0.16.
The cost for generation of logical form from a prompt with 5 in-context examples is \$0.16. 
Thus, the approximate cost for inference of one question by \sysold{}-U is \$0.32. 

The cost for generation of logical form from a prompt with 0 in-context examples is \$0.04.
The cost of checking whether two natural language questions are equivalent or not, using few-shot exemplars and chain of thought prompting is also \$0.04. 
The approximate cost of inference of one question by FUn-FuSIC varies between \$0.24 and \$0.48. The average cost over 50 randomly sampled questions from the test set is around \$0.34. 

Hence, the two models are comparable in terms of cost. 

\subsection{Model Adaptation Details}

Here we discuss adaptation details for the models that we have built upon (\sysold{}), used as retrievers (RetinaQA) and for comparison.

\subsubsection{\sysold{} Details} \label{fusic_high_level}
Our proposed approach \sys{} builds upon the the base architecture of \sysold{}~\cite{patidar-etal-2024-fusic}. \sysold{} has a three step pipeline: (a) Supervised Retrieval: a supervised retriever, trained on the source domain and optionally fine-tuned on the target domain is used to obtain the top-100 classes, relations and paths that are relevant to the question asked, 
(c) LLM Generation: We provide the top-10 classes, top-10 relations and top-5 paths along with few-shot exemplars to generate the SPARQL query. 

Since no code is available for this model, we use our own implementation based on the description in the paper.
For \sysold{}, and \sys{} we use LLM $temperature=0$.

\subsubsection{Training Details for Supervised Models} \label{comp_and_infra}

We use Hugging Face ~\cite{wolf-etal-2020-transformers}, PyTorch \cite{pytorch} for our experiments and use the Freebase setup specified on github \footnote{\url{https://github.com/dki-lab/Freebase-Setup}} . We use NVIDIA A100 GPU with 40 GB GPU memory and 32 GB RAM. For training the discriminator module of RetinaQA, we require 2 GPUs. 
(1) For the answerable experiments, we use the supervised models as specified in \cite{patidar-etal-2024-fusic}. (2) For the unanswerability experiments, we train all models from scratch. (a) We use RnG-KBQA entity linker \footnote{\url{https://github.com/salesforce/rng-kbqa/tree/main/GrailQA/entity\_linker}} (BSD 3-Clause License) trained on the answerable subset of GrailQAbility for all our experiments. (b) We train the RnG-KBQA path retriever on answerable subset of WebQSP\footnote{\url{https://github.com/salesforce/rng-kbqa/blob/main/WebQSP/scripts/run\_ranker.sh}} (BSD 3-Clause License). The number of training epochs is determined by the performance of the model over the answerable questions in the dev set. (c) We train the TIARA schema retriever on the answerable subset of WebQSP \footnote{\url{https://github.com/microsoft/KC/tree/main/papers/TIARA/src}} (MIT License) (d) We train the sketch generator and discriminator of RetinaQA on the answerable subset of WebQSP\footnote{\url{https://github.com/dair-iitd/RetinaQA}}.  

\subsubsection{Inference Details for Supervised Models}
\label{inference_details}
We train all RetinaQA components on the source WebQSP's training set, using the corresponding target domain's dev set as a validation set for early stopping.
In the absence of unanswerable questions for training, both models use a threshold fine-tuned on a dev set to detect schema-level unanswerability. 
We again use the target dev sets for this.

We use the dev set in RetinaQA, during discriminator inference for different purposes. (A) Determining how to best utilize the candidate paths. The possibilities are (i) not providing candidate paths, (ii) providing candidate paths in GrailQA format, and (iii) providing candidate paths in WebQSP format. We select the best alternative based upon the performance of the model over the dev set. For the WebQSP $\rightarrow$ GrailQAbility dataset, we observe (ii) works best, whereas for the WebQSP $\rightarrow$ GraphQAbility dataset, we observe (i) works best. (B) Determining the threshold value. RetinaQA applies a threshold on the scores - for a question, if the highest score candidate logical form has a score less than the threshold, the question is labeled as NK. We choose the optimal value of the threshold to maximize the overall EM-s score over the dev set.

\subsection{Pangu Adaptation Details}
Similar to RetinaQA, we train all Pangu components on WebQSP, using the corresponding target domain's dev set as a validation set for early stopping. We use one GPU for training. Same as RetinaQA, we use the dev set to determine the threshold for schema-level unanswerability. Pangu-T applies a threshold on the scores - for a question, if the highest score candidate logical form has a score less than the threshold, the question is labeled as NK. We choose the optimal value of the threshold to maximize the overall EM-s score over the dev set. 

\subsubsection{KB-Binder Adaptation Details}
\label{subsec:appendix_kbbinder}
For KB-Binder, we make use of publicly available code \footnote{\url{https://github.com/ltl3A87/KB-BINDER}} (MIT License).
We use self-consistency and majority voting with 6 examples, as in the experiments in the paper. 
In the retrieval(-R) setting, KB-Binder samples demonstration examples by retrieving from the entire available training data.
We restrict its retrieval to our target training set $D_t$ with 25 examples. 
KB-Binder reports experiments using \texttt{code-davinci-002} as the LLM.
For consistency and fair comparison, we replace this with \texttt{gpt-4-0613} as in other LLM-equipped models.
KB-Binder generates logical forms in s-expression, which we preserve.

\begin{table}[ht]
\begin{center}
    \small
    \centering
    \begin{tabular}{l|c|c}
        
        \hline
        & {\bf WebQSP} $\rightarrow$  & {\bf WebQSP}  $\rightarrow$\\ 
        {\bf Model}  & {\bf GrailQAbility} & {\bf GraphQAbility} \\
        \hline
        V3(a) & 92 & 96 \\ 
        V3(b) & 94 & 90\\ 
        V3(c) & 92 & 94\\ 
        V3(overall) & 88 & 90\\
        V4(b) & 75 & 68\\
        \hline
    \end{tabular}
    \caption{Accuracy of weak verifier on the two datasets by manual analysis of 100 instances from the test sets.}
    \label{tab:verifier_accuracy_detail}
\end{center}
\end{table}

\subsection{Accuracy of Weak Verifiers}\label{appendix:subsec:weak-verifiers}

Accuracy of the two weak verifiers are recorded in detail in Tab.~\ref{tab:verifier_accuracy_detail}.

\subsection{\sys{} Prompts}
Here we provide details of various prompts used by \sys{}.

\subsubsection{PUn prompt} \label{pun_prompt}
The following prompt is for Prompting for Unanswerability (PUn).

\begin{tcolorbox}[colback=gray!10!white,colframe=black!75!black,title=Header Prompt]
\texttt{Translate the following question to sparql for Freebase based on the candidate sparql, candidate entities, candidate relations and candidate entity types which are separated by "|" respectively. Please do not include any other relations, entities and entity types. Your final sparql can have three scenarios: 1. When you need to just pick from candidate sparql. 2. When you need to extend one of candidate sparql using the candidate relations and entity types. 3. When you will generate a new sparql only using the candidate entities, relations and entity types. For  entity type check please use this relation "type.object.type".D o not use entity names in the query. Use specified mids. If it is impossible to construct a query using the provided candidate relations or types, return "NK". Make sure that the original question can be regenerated only using the identified entity types, specific entities and relations.}
\end{tcolorbox}
\begin{tcolorbox}[colback=gray!10!white,colframe=black!75!black,title=NK exemplar]
\texttt{\textbf{Question:} the tv episode segments spam fall under what subject? \textbf{Candidate entities:}  spam m.04vbm Candidate paths: SELECT DISTINCT ?xWHERE {?x0 ns:tv.tv\_segment\_performance.segment ns:m.04vbm .?x0 ns:tv.tv\_segment\_performance.segment ?x .?x ns:type.object.type ns:tv.tv\_episode\_segment .} | ...
\textbf{Candidate entity types: }tv.tv\_series\_episode| tv.tv\_episode\_segment | ...
\textbf{Candidate relations: }tv.tv\_series\_episode.segments (type:tv.tv\_series\_episode R type:tv.tv\_episode\_segment)| tv.tv\_subject.tv\_programs (type:tv.tv\_subject R type:tv.tv\_program)|...
\textbf{sparql:}NK}
\end{tcolorbox}
\begin{tcolorbox}[colback=gray!10!white,colframe=black!75!black,title=Question Prompt]
\texttt{\textbf{Question:} which school newspaper deals with the same subject as the onion? 
\textbf{Candidate entities:}  the onion m.0hpsvmv 
\textbf{Candidate paths:} SELECT DISTINCT ?xWHERE {ns:m.0hpsvmv ns:book.newspaper.circulation\_areas ?x0 .?x0 ns:periodicals.newspapers ?x .?x ns:type.object.type ns:book.newspaper .} |...
\textbf{Candidate entity types:} education.school\_newspaper| type:book.newspaper...
\textbf{Candidate relations:} education.school\_newspaper.school (type:education.school\_newspaper R type:education.educational\_institution) | book.newspaper\_issue.newspaper (type:book.newspaper\_issue R type:book.newspaper)|...
\textbf{sparql:}}
\end{tcolorbox}

\subsubsection{FUN Prompts} 

The following is the prompt used by FUn for accommodating feedback from verifier V1.

\label{fun_prompt}
\begin{tcolorbox}[colback=red!5!white,colframe=red!75!black,title=Syntax error(V1) Feedback]
\texttt{
Correct the syntax of the following sparql query. Return ONLY the corrected sparql query without any explanation
\textbf{sparql: } SELECT ?x AND ?y {...}
\textbf{Virtuoso error: } word AND not defined 
}
\end{tcolorbox}

The following is the prompt used by FUn for accommodating feedback from verifier V2.

\begin{tcolorbox}[colback=red!5!white,colframe=red!75!black,title=KB Inconsistency(V2) Feedback]
\texttt{The generated sparql has a semantic issue warning:  The types of relations don't match for variable ?x in the query. The assigned relation types by ['computer.computer\_emulator.computer', 'type.object.type computer.computer\_peripheral'] are ['computer.computer', 'computer.computer\_peripheral']. These types are mutually incompatible... Please generate again a different executable sparql using the same context and constraints. DO NOT APOLOGIZE - just return the best you can try.
}
\end{tcolorbox}

The following is the prompt used by FUn for accommodating feedback from verifier V3.

\begin{tcolorbox}[colback=red!5!white,colframe=red!75!black,title=Question Logical form disagreement(V3) feedback]
\texttt{
The question that you answer is NOT same as what you've been asked for! You have answered the question "Which opera productions has Gino Marinuzzi conducted?" but you were asked to answer "what is the name of the premiere opera production conducted by gino marinuzzi?". Please generate again a different executable sparql using the relations, classes and entities provided earlier. DO NOT APOLOGIZE - just return the best you can try.
}
\end{tcolorbox}

The following three prompts are used by FUn for accommodating feedback from verifier V4.

\begin{tcolorbox}[colback=red!5!white,colframe=red!75!black,title=Answer Inconsistency(V4b) feedback]
\texttt{
The generated sparql gives an empty answer when executed on freebase KG, Please generate again a different executable sparql using the same context and constraints.
}
\end{tcolorbox}

\begin{tcolorbox}[colback=red!5!white,colframe=red!75!black,title=Intermediate Node(V4a) feedback]
\texttt{
The generated sparql returns an intermediate type node when executed on the freebase KG. Maybe the answer node is an adjacent node to what we currently query for. Please generate again a different executable sparql using the same context and constraints.
}
\end{tcolorbox}

\begin{tcolorbox}[colback=red!5!white,colframe=red!75!black,title=Answer Inconsistency(V4a) feedback]
\texttt{
The logical form upon execution returns International System of Units, which is not answering the question. Please reconstruct the query using same context and constraints. 
}
\end{tcolorbox}

\subsubsection{Prompt for Question Logical Form Agreement Verifier (V3) } \label{qlf_prompt}
The few shots provided for verifying question logical form agreement are derived from $D^t$. 
We obtain positive samples from the dataset $D^t$ directly, using the questions and gold logical forms. 
For obtaining negative samples, we perform zero-shot \sysold{} inference over $D^t$. Then we consider those questions for which the predicted logical form is different from the gold logical form. 

First, we perform back-translation to obtain natural language question from the logical form using the following prompt.

\begin{tcolorbox}[colback=blue!5!white,colframe=blue!75!black,title=Naturalization of variable names(V3(i))]
\texttt{
change the sparql query to have variable names representative of what objects they refer to. transform the variable names in this query. Do NOT change the prefix headers and relation names 
}
\end{tcolorbox}

\begin{tcolorbox}[colback=blue!5!white,colframe=blue!75!black,title=Conversion of Logical Form into Natural Language Question(V3(ii))]

\texttt{
Convert this sparql query into a natural language question. Make the question as natural as possible. SELECT DISTINCT ?unfinishedWork WHERE $\{$ Le Moulin de Blute-Fin ns:media.unfinished\_work ?unfinishedWork . ?unfinishedWork ns:type.object.type ns:media.unfinished\_work . \}
}
\end{tcolorbox}

We use few-shot LLM prompting to obtain the explanation for question and logical form agreement or disagreement. 

\begin{tcolorbox}[colback=blue!5!white,colframe=blue!75!black,title=Explanation Generation Prompt]
\texttt{
\textbf{Explain why the two questions are different}.
\textbf{Question we answer:} who all like to eat apple or mango?
\textbf{Question originally asked:} what are the people who enjoy both apple and mango?
\textbf{explanation:} The question we answer returns people. The question originally asked also returns people. 
The question we answer finds those people who like eating apple, those people who like eating apple. The  question originally asked also finds those people who like eating apple, those people who like eating apple.
The question we answer uses logical operator OR. However, the question originally asked uses the logical operator AND  
Hence, they are different. 
\textbf{[total 3 exemplars]}
\textbf{Question we answer}: Which game engines are successors to the Unreal Engine?
\textbf{Question originally asked:} which video game engine's successor engine is unreal engine?
\textbf{explanation: }
}
\end{tcolorbox}

These few-shots for obtaining the explanation are dataset independent, and are manually written. 
Some examples of few shots are  below.

\begin{tcolorbox}[colback=blue!5!white,colframe=blue!75!black,title=Question Logical Form Agreement Check(V3(iii))]
\texttt{
\textbf{Question we answer:} Who are the cricket players who have made exactly 31 stumps in one day internationals?
\textbf{Question originally asked:} name the cricket player who has 31 odi stumps.
\textbf{explanation:} The question we answer returns cricket players. The question originally asked also returns cricket players.
The question we answer finds cricket players who have made exactly 31 stumps in one day internationals. The question originally asked also finds cricket players who have made 31 stumps in one day internationals.
Both questions involve no mathematical or logical operators.
\textbf{Hence, they are same.}
\textbf{Question we answer:} Which game engines are successors to the Unreal Engine?
\textbf{Question originally asked:} which video game engine's successor engine is unreal engine?
\textbf{explanation:} The question we answer returns game engines. The question originally asked also returns game engines. 
The question we answer finds successors to the Unreal Engine. The question originally asked finds the predecessor of the Unreal Engine. 
The reasoning steps followed by the two questions are different. 
\textbf{Hence, they are different.} 
\textbf{[total 6 exemplars]}
\textbf{Question we answer: } Which cars drive at a speed of 80?
\textbf{Question originally asked: }name the car with driving speed at least 80?
\textbf{explanation: }
}
\end{tcolorbox}

\subsubsection{scUn Prompt for Logical Form Selection}
We store the back-translated natural language questions along with each predicted logical form. We use the following prompt to select the closest back-translated natural language question to the original question. 

\begin{tcolorbox}[colback=green!5!white,colframe=green!75!black,title=Best Logical Form Selection Prompt]
\texttt{
\textbf{orig\_nl\_qn =} which surf films has sarah finn served as the casting director?
1. \textbf{pred\_nl:} Which surfing films has Sarah Finn directed the casting for?
2. \textbf{pred\_nl:} Which surfing films has Sarah Finn been the casting director for?
of the 2 predicted nl questions, which is closest to the original nl question. Even if none is very close, return the one that is semantically closest? Please explain your answer as well
}
\end{tcolorbox}

\subsection{Example of FUn Iterations}\label{subsec_fun_real}

The following shows 3 Fun iterations for an example question from the test set of WebQSP $\rightarrow$ GrailQAbility dataset.

\begin{CJK}{UTF8}{mj}
\textbf{Question:} what is the musical genre of the recording who m i (feat. 일리닛, new champ, myk)?\\
Entities:  who m i (feat. 일리닛, new champ, myk) m.0123lk0s
\end{CJK}
\begin{tcolorbox}[colback=gray!10!white,colframe=black!75!black,title=Logical Form 1]
\texttt{?x \{\\
    \ \ \ \ :m.0123lk0s :music.genre.recordings ?x\\
    \ \ \ \ ?x :type.object.type :music.genre\}}
\end{tcolorbox}

\begin{tcolorbox}[colback=red!5!white,colframe=red!75!black,title=Feedback 1]
\textbf{Type:} Logical Form Semantic Error \\
\textbf{Description:} The types of relations don't match for entity in the query. The assigned relation types by ['music.genre.recordings'] are ['music.genre']. These types are not associated with this entity in the KB.
\end{tcolorbox}

\begin{tcolorbox}[colback=gray!10!white,colframe=black!75!black,title=Logical Form 2]
\texttt{?y \{\\
    \ \ \ \ :m.0123lk0s :music.recording.artist ?x\\
    \ \ \ \ ?x :music.artist.genre ?y\\
    \ \ \ \ ?y :type.object.type :music.genre\}}
\end{tcolorbox}

\begin{tcolorbox}[colback=red!5!white,colframe=red!75!black,title=Feedback 2]
\textbf{Type:} Natural Language Semantic Error \\
\textbf{Description:} The question originally asked genre of the song. However, the question we answer returns genre associated with artist of the song.
\end{tcolorbox}

\begin{tcolorbox}[colback=gray!10!white,colframe=black!75!black,title=Logical Form 3]
\texttt{?x \{ ?x \\
    \ \ \ \ :music.genre.recordings :m.0123lk0s\\
    \ \ \ \ ?x :type.object.type :music.genre\}
    }
\end{tcolorbox}

\begin{tcolorbox}[colback=green!5!white,colframe=green!65!black,title=Feedback 3]
\textbf{Status:} Passes All Checks \\
\textbf{Details:} lf3 satisfies syntax, logical form semantic, natural language semantic and execution checks.
\end{tcolorbox}

\end{document}